\documentclass[conference]{IEEEtran}
\IEEEoverridecommandlockouts
\usepackage{cite}
\usepackage{amsmath,amssymb,amsfonts}
\usepackage{mathrsfs}
\usepackage{algorithmic}
\usepackage{algorithm}
\usepackage{graphicx}
\usepackage{textcomp}
\usepackage{xcolor}
\usepackage{url}
\usepackage{balance}
\usepackage{tabularx}
\usepackage{multirow}
\usepackage{booktabs}
\usepackage{natbib}
\def\BibTeX{{\rm B\kern-.05em{\sc i\kern-.025em b}\kern-.08em
    T\kern-.1667em\lower.7ex\hbox{E}\kern-.125emX}}
\newcommand{\emp}[1]{\emph{#1}}
\begin{document}

\title{LLM-VLM Fusion Framework for Autonomous Maritime Port Inspection using a
Heterogeneous UAV-USV System
\thanks{$^{1}$ Khalifa University Center for Autonomous Robotic Systems (KUCARS), Khalifa University, United Arab Emirates.}%
 
\thanks{$^{*}$ This publication is based upon work supported by the Khalifa University of Science and Technology under Award No. RC1-2018-KUCARS.\\ 
 Corresponding Author, Email: irfan.hussain@ku.ac.ae}
}

\author{Muhayy Ud Din$^{1}$, Waseem Akram$^{1}$,  Ahsan B. Bakht$^{1}$, and Irfan Hussain$^{1,*}$

}

\maketitle

\begin{abstract}
Maritime port inspection plays a critical role in ensuring safety, regulatory compliance, and operational efficiency in complex maritime environments. However, existing inspection methods often rely on manual operations and conventional computer vision techniques that lack scalability and contextual understanding. This study introduces a novel integrated engineering framework that utilizes the synergy between Large Language Models (LLMs) and Vision Language Models (VLMs) to enable autonomous maritime port inspection using cooperative aerial and surface robotic platforms. The proposed framework replaces traditional state-machine mission planners with LLM-driven symbolic planning and improved perception pipelines through VLM-based semantic inspection, enabling context-aware and adaptive monitoring. The LLM module translates natural language mission instructions into executable symbolic plans with dependency graphs that encode operational constraints and ensure safe UAV-USV coordination.
Meanwhile, the VLM module performs real-time semantic inspection and compliance assessment, generating structured reports with contextual reasoning. The framework was validated using the extended MBZIRC Maritime Simulator with realistic port infrastructure and further assessed through real-world robotic inspection trials. The lightweight on-board design ensures suitability for resource-constrained maritime platforms, advancing the development of intelligent, autonomous inspection systems.
Project resources (code and videos) can be found here: \hbox{\textcolor{blue}{\url{https://github.com/Muhayyuddin/llm-vlm-fusion-port-inspection}}}.
\end{abstract}

\date{}

\maketitle

\section{Introduction}

Seaports play a crucial role in international trade, as more than 80\% of the volume of international trade in goods is carried by sea~\cite{unctad2025shipping}. These maritime ports require a consistent evaluation of their infrastructure to ensure operational efficiency, safety, and compliance with environmental and security regulations~\cite{statheros2008autonomous, noel2019autonomous}. Traditional inspection techniques typically depend on manual visual inspections or human-operated boats, which are often time-consuming and expensive. It includes substantial safety risks to personnel operating in hazardous industrial port environments with heavy machinery and high-traffic vessel movements. Performing comprehensive inspections regularly across large port regions is a challenge. In addition, high operational costs are associated with the deployment of specialized inspection personnel. Recent works such as ~\cite{tian2024seaair,peti2023search} highlighted the crucial need for robust and automated inspection methods that can use emerging technologies such as autonomous unmanned systems to address the scalability and efficiency challenges inherent in the inspection of maritime infrastructure.


\begin{figure}[t]
    \centering
    \includegraphics[width=\linewidth]{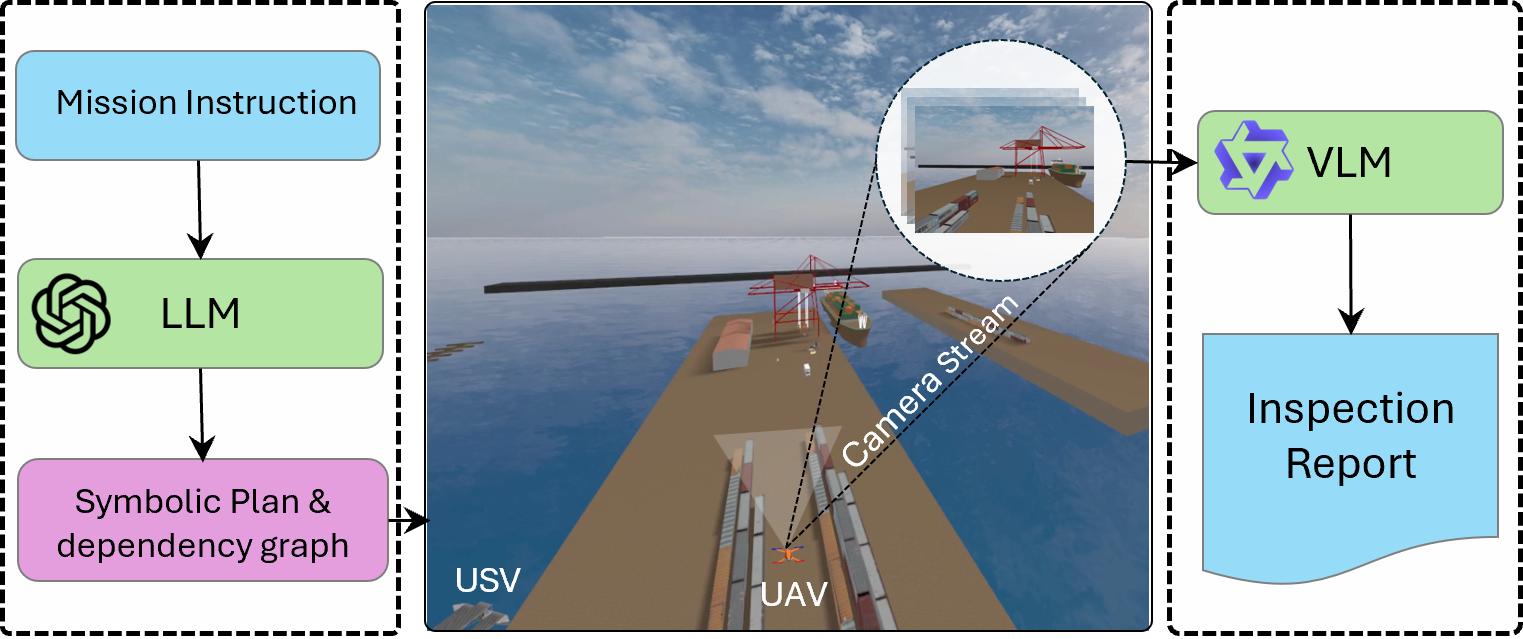}
    \caption{LLM-VLM framework for autonomous maritime inspection. 
Mission instructions are processed by a LLM to generate symbolic plans and 
dependency graphs for UAV-USV coordination. The VLM then analyzes sensor 
data from the inspection scene to produce structured inspection reports, enabling semantic understanding, anomaly detection, and compliance assessment in dynamic port environments.}

    \label{fig:types_of_disturbances}
\end{figure}

The heterogeneous UAV-USV systems show significant potential for maritime operations, providing essential capabilities that single-platform systems cannot achieve. These combined systems play a crucial role in the current maritime security, environmental monitoring, and inspection tasks, where the combination of UAV and USV platforms creates significant operational advantages \cite{tian2024seaair}\cite{peti2023search}. Maritime environments present challenges that require the collaborative capabilities of heterogeneous systems. USVs provide stable platforms for extended operations and heavy payload deployment with autonomous navigation capabilities through busy waters, close-up inspection of above-waterline structures, and floating assets, while UAVs offer rapid deployment, high-altitude observation points, and access to areas that are not reachable by surface vessels \cite{kim2023survey}. Such unmanned systems have the potential to complete maritime tasks that are unattainable for manned vehicles, significantly reduce human exposure to risk, and provide cost-effective solutions for large-scale maritime operations \cite{kim2023survey}\cite{ahmed2024comprehensive}.


The current state-of-the-art in heterogeneous UAV-USV systems focuses primarily on addressing fundamental technical challenges related to robust platform coordination, such as achieving reliable UAV landing on moving USV platforms \cite{tian2024seaair,stephenson2024time,prochazka2024trajectory}, developing advanced control systems for robust navigation in the presence of environmental disturbances, and advance localization strategies in the GNSS denied environment \cite{jarraya2025gnssdenied,zhou2025satloc}. 
For mission planning, UAV-USV heterogeneous systems primarily rely on fixed mission planning architectures based on predefined state machines and rigid operational frameworks, which limit their adaptability and autonomous decision-making capabilities. Traditional mission planning approaches utilize preconfigured plans that cannot cope with the entire complexity of maritime operations, and manually provided schedules that require extensive human intervention \cite{evers2019automated}. The execution of a mission follows predetermined sequences of actions without the ability to adapt to dynamic environmental changes or unexpected scenarios \cite{sorensen2025towards}.

For inspection tasks, existing systems are based on classical computer vision pipelines for object detection, particularly traditional object detection algorithms such as YOLO-based approaches and conventional feature matching methods that focus primarily on detecting predefined categories \cite{leira2021object, wang2022maritime}.  The reliance on traditional computer vision frameworks fundamentally constrains the inspection capabilities of heterogeneous systems, as they cannot provide semantic understanding of complex maritime scenes, they may not adapt to diverse operational contexts, and lack the contextual reasoning necessary for advanced autonomous inspection tasks \cite{zhang2024comprehensive}. 


To address these gaps, we propose a novel approach that uses Large Language Models (LLMs) for intelligent mission planning for unmanned heterogeneous systems and vision language models (VLMs) for advanced inspection capabilities. Our approach uses LLM such as GPT-4 and similar architectures, to enable dynamic mission planning through natural language instruction, allowing operators to specify complex maritime missions using intuitive text commands that are automatically translated into executable mission plan for USV and UAV \cite{maritime2025mission}. For enhanced inspection capabilities, we employ VLMs such as Florence-2 that fuse visual and
textual modalities, allowing inspection systems that provide sophisticated scene understanding and contextual analysis beyond traditional computer vision approaches, allowing the system to interpret complex maritime scenarios and make intelligent decisions based on multimodal environmental understanding. 

To our knowledge, this is the first study that applies LLM-VLM fusion approach for maritime port inspection using a heterogeneous UAV-USV system. The key contributions of this work is the LLM-VLM fusion framework for autonomous port inspection using the UAV-USV heterogeneous system. The main contribution is followed by the following contributions.
\\
\begin{itemize}
    \item \textit{LLM-Driven Symbolic Mission Planning with Integrated Preconditions:} Development of a novel framework that converts natural language inspection  instructions into executable symbolic plans for heterogeneous UAV-USV systems. The LLM generates both the symbolic action sequences and their associated preconditions, incorporating maritime safety knowledge, operational constraints, and inter-platform dependencies within a unified planning output.
    
    \item \textit{Dependency Graph-Based Execution Management:} Design and implementation of a communication manager that constructs dependency graphs from LLM-generated symbolic plans and preconditions, enabling automatic resolution of task dependencies and safe parallel execution of UAV-USV subtasks under dynamic maritime conditions. This ensures that all preconditions are satisfied before task execution, while maximizing efficiency through intelligent scheduling.
    
    \item \textit{VLM-Based Semantic Inspection System:} Implementation of lightweight VLMs for real-time multimodal inspection, enabling context-aware anomaly detection, regulatory compliance checking, and structured report generation that goes beyond traditional computer vision pipelines through semantic reasoning and natural language understanding.
    
    \item \textit{Comprehensive AI Model Benchmarking:} Systematic evaluation of multiple LLM (GPT-4o, GPT-3.5-Turbo, GPT-4, Gemini, LLaMA) for mission planning accuracy and execution success, and lightweight VLM (SmolVLM, Florence-2, Qwen2-VL, Moondream2, GIT-base) for inspection accuracy, analysis of response time, computational efficiency, and deployment feasibility for resource-constrained maritime platforms.
\end{itemize}

\section{Related Work}
This section provides an overview of the current state of the art in three key areas relevant to this study: (i) heterogeneous UAV-USV systems, which integrate aerial and surface platforms for cooperative maritime operations; (ii) mission planning approaches, with a particular focus on recent advances using LLM for adaptive and natural language driven decision making; and (iii) perception methods for maritime inspection, where VLM are emerging as powerful tools for semantic scene understanding and robust inspection in dynamic environments.
\subsection{UAV-USV Heterogeneous Systems}%
Heterogeneous UAV-USV systems have gained significant interest in port and maritime evaluation through the integration of aerial surveillance with surface navigation capability. The first collaborative UAV-USV systems that have been practically implemented include a 3D printed UAV capable of landing guided by IR beacons on a catamaran USV, which improves the detection range and precision in variable lighting conditions \cite{niu2021design,akram2024enhancing}. In the domain of coordinated control, leader-follower architectures and finite-time fuzzy sliding mode control have been explored to maintain formation stability in UAV-USV systems, with robust performance demonstrated in simulation environments under limited communication bandwidth \cite{xue2021distributed}. In addition, USVs equipped with multisensor suites have facilitated effective hydrographic surveys, allowing the mapping of harbor areas adjacent to dock structures, which is essential to maintain port stability and safety \cite{specht2024hydrographic}. 

Recent real-world deployments have further demonstrated the adaptability of cooperative UAV-USV systems \cite{akram2024long}. One notable example is the USV drone carrier, which is an electric catamaran equipped with multiple sensors and a robotic manipulator, which allowed fully autonomous UAV launch and landing for object manipulation in GNSS denied environments~
\cite{dong2025dronecarrier}. Moreover, collaborative guidance has also allowed UAVs to track and land on dynamically moving USV decks in high-wave environments, using six-degree-of-freedom (6-DOF) USV models, multisensor fusion, and predictive trajectory planning \cite{novak2024towards}. In addition, tethered UAV–USV configurations have been explored for cooperative object manipulation on water surfaces, where models predictive control (MPC)-based strategies demonstrated precision and faster disturbance recovery in simulation \cite{novak2024tethered}. Deep learning-driven visual navigation, coupled with reinforcement learning USV control, has also shown potential to enhance situational awareness and disturbance rejection in collaborative maritime search and rescue operations \cite{cooperative2022visual}. Together, these developments underscore the emergence of heterogeneous UAV-USV systems as highly capable, adaptive, and robust options for autonomous port and maritime inspection.

\subsection{Maritime Mission Planning with LLMs}

Maritime UAV-USV mission planning has typically employed a rule-based system, state machine behaviors, and predefined route planners that implement search, approach, inspection, and recovery as fixed transitions or waypoint sequences \cite{xing2023review,chu2024evolution,hashali2024route,zhang2024usvuav,zhao2024oe,hinostroza2024temporal}. 
These architectures face challenges when dealing with complex and dynamic environments, such as dynamic sea conditions. They are also not well equipped to handle multiple objectives simultaneously, such as ensuring safety, saving time, and conserving energy. In addition, these systems struggle with planning and coordination over long time frames. These issues often require substantial manual adjustments to function properly, and systems may not perform well outside their expected operational settings and may lead to unreliable behavior.
\cite{xing2023review,chu2024evolution,wang2025riskaware}.

Large language models have recently gained attention in the field of maritime operations. LLMs have been used to convert natural language instructions into executable plans for different maritime tasks, such as piloting Autonomous Underwater Vehicles \cite{yang2024oceanplan,yang2023oceanchat}, programming missions with subsea deployments\cite{chen2025word2wave}, decision-making considering maritime regulations for USVs \cite{agyei2024colregs} , and maneuvering surface vehicles\cite{li2025visionllm}. These applications illustrate the versatility of LLMs in generating maritime plans from textual input. 
Some of the recent studies explore the potential of LLMs in improving mission planning for maritime robotic systems. For example, \cite{maritime2025mission} introduced a framework that uses GPT4 to translate high-level natural language commands into symbolic mission plans for USVs, with integrated feedback from low-level controllers allowing real-time adaptation to dynamic maritime environments. Similarly, for underwater scenarios approaches like OceanPlan, Aquachat are proposed that use hierarchical LLM-based planner for underwater vehicles for inspection missions\cite{yang2024oceanplan,akram2025aquachat,saad2025aquachat++}. 

None of these methodologies employ LLM as a high-level mission planner for heterogeneous systems. In this role, the LLM would be tasked with creating an overarching symbolic plan for the heterogeneous system by deciding on navigation strategies, inspection strategies, and determining whether to utilize the UAV, USV, or both for the mission.

\subsection{VLMs for Maritime Perception}%
Traditional vision systems for maritime inspection and monitoring applications rely on sequential processing pipelines that integrate multiple specialized components to analyze various maritime assets, including ship hulls, storage tanks, offshore structures, vessels, and port infrastructure \cite{zhang2021survey,patel2022ships,zhao2024ships}. These conventional pipelines typically begin with image enhancement techniques, such as dehazing algorithms and glare or motion compensation methods, to improve image quality in challenging maritime conditions \cite{yasir2023sarslr,zhang2024fowt}. The enhanced images are then processed through feature extraction modules followed by task-specific detection and segmentation algorithms designed to identify ships, structural cracks, marine growth, and various infrastructure defects \cite{andersen2024remote,pena2024uavtank,lv2025usv}. 
Although these conventional vision systems have made substantial progress by incorporating deep learning techniques, they still face several fundamental limitations\cite{yasir2023sarslr,zhao2024ships,andersen2024remote}.
The systems show inconsistency in dealing with domain-shift situations, such as changes in weather conditions, changes in object scale and orientation. Furthermore, they have restricted ability in recognizing unseen objects, indicating difficulty in identifying object categories that weren't part of their initial training dataset. The development and deployment of these systems also require extensive and costly dataset preparation processes whenever new asset types or inspection scenarios are introduced.

VLMs offer potential solutions to these limitations by creating aligned representations between visual content and textual descriptions, enabling advanced capabilities such as open-vocabulary detection, visual grounding, image captioning, and semantic segmentation \cite{li2023blip2,xiao2024florence2}. General purpose VLMs, including Florence-2 and Qwen2-VL provide zero-shot and few-shot transfer learning capabilities through natural language text prompts \cite{minderer2023owlvit,liu2024groundingdino}. Remote sensing adaptations of these technologies, such as RemoteCLIP, demonstrate effective text-prompted detection of novel object categories in aerial and maritime environments \cite{rao2023remoteclip,pan2024laedino}. Recent advances in training-free and open-vocabulary segmentation methods have further demonstrated the feasibility of achieving pixel-level labeling without requiring class-specific supervised training data \cite{li2025segearthov}.

Although emerging maritime-focused VLMs demonstrate initial promise in unifying multisource sensing capabilities that integrate diverse data types such as optical imagery for tasks such as ship detection, tracking, and reidentification \cite{li2024popeye}, the application of VLMs specifically for inspecting maritime infrastructure is still largely unexplored \cite{akram2025review}. Despite the theoretical advantages of VLMs, very few maritime inspection systems currently utilize these models for practical maritime inspection and surveillance tasks. However, these early developments suggest potential pathways toward promptable inspection systems and enhanced situational awareness capabilities that could eventually lead to improved robustness to environmental variations and diversity in maritime monitoring applications.

This study addresses the identified gap in the current state-of-the-art by introducing a novel heterogeneous maritime inspection framework that integrates LLM for multi-agent mission planning and VLM for enhanced inspection capabilities. The proposed framework utilizes VLMs to provide advanced semantic understanding of maritime environments, utilizing pretrained lightweight VLM architectures that are specifically designed for deployment on resource-constrained onboard systems. This approach enables dynamic mission coordination through natural language instructions while providing robust inspection capabilities that can adapt to diverse maritime scenarios without requiring extensive retraining or manual parameter adjustment.

\begin{figure*}[t]
    \centering
    \includegraphics[width=\linewidth]{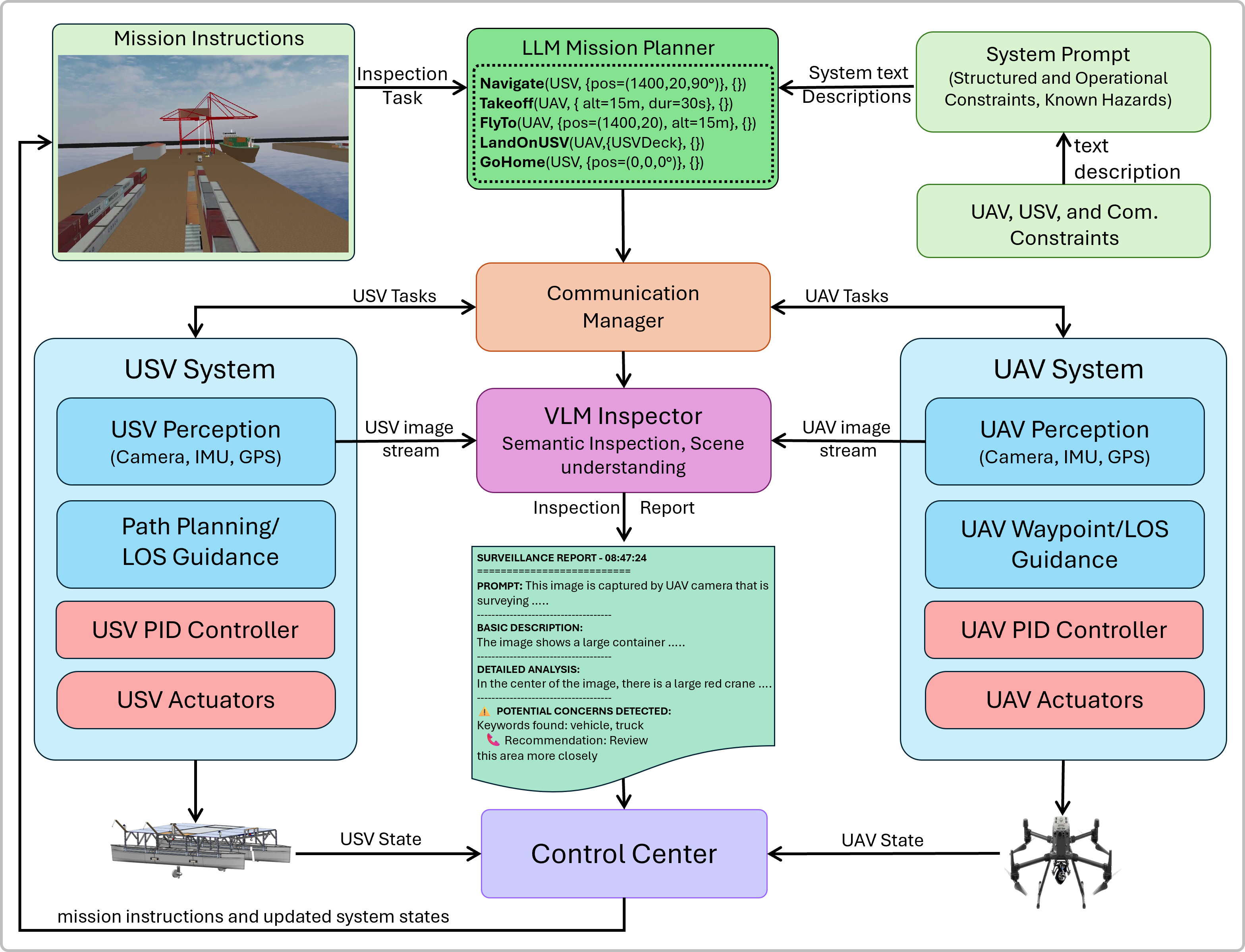}
    \caption{Overview of the proposed LLM-VLM fusion framework for autonomous maritime inspection. 
Mission instructions and system prompts, including operational constraints and hazard knowledge, 
are processed by the LLM mission planner to generate symbolic plans and dependency graphs for 
safe UAV-USV coordination. The USV and UAV subsystems perform perception, waypoint guidance, 
and control, while the VLM Inspector provides semantic scene understanding, anomaly detection, 
and compliance assessment. The communication manager synchronizes information flow across 
platforms, and structured inspection reports are transmitted to the control center for operator review.}

    \label{fig:framework}
\end{figure*}
\section{System Overview and Problem Formulation}

\subsection{Maritime Environment and Platform Overview}
The maritime port inspection problem involves the systematic monitoring and assessment of port infrastructure, vessel compliance, and operational activities using autonomous robotic systems equipped with intelligent perception capabilities. The port environment is represented as a bounded domain $\mathcal{P} \subset \mathbb{R}^3$, which contains static obstacles $\mathcal{O}_s = \{o^s_1, o^s_2, \ldots, o^s_n\}$ (such as cranes, buildings and docked vessels) and dynamic obstacles $\mathcal{O}_d = \{o^d_1(t), o^d_2(t), \ldots, o^d_m(t)\}$ (including moving vessels), where $t$ represents time. A heterogeneous robotic system 
\(\mathcal{R} = \{\mathcal{R}_{\text{USV}}, \mathcal{R}_{\text{UAV}}\}\), 
comprising a USV and a UAV, is used to perform autonomous port inspection tasks.

The USV serves as the primary coordination platform and provides a mobile base for UAV operations. Its kinematic state is defined as:
\begin{equation}
\mathbf{x}_s = [x_s, y_s, \psi_s, v_s]^\top
\end{equation}
where $(x_s, y_s)$ represents the position in the horizontal plane, $\psi_s$ is the yaw angle, and $v_s$ is the surge speed. Assuming negligible sway and heave motions for high-level planning, the USV follows a differential drive kinematic model:
\begin{align}
\dot{x}_s &= v_s\cos\psi_s, &
\dot{y}_s &= v_s\sin\psi_s, \nonumber \\
\dot{\psi}_s &= r_s, &
\dot{v}_s &= a_s,
\label{eq:usv-kin}
\end{align}
with control inputs $\mathbf{u}_s = [a_s, r_s]^\top$, where $a_s$ is the surge acceleration command and $r_s$ is the yaw rate command.

The UAV provides aerial surveillance and detailed inspection capabilities.  The UAV state is 
$
\mathbf{x}_a = [x_a, y_a, z_a, \psi_a]^\top
$
. For simplicity, we assume that there are no movements along the roll and pitch axes for high-level planning. We use a point-mass model:
\begin{align}
\ddot{\mathbf{p}}_a &= \mathbf{u}_a, \quad \mathbf{p}_a = [x_a, y_a, z_a]^\top, \quad \mathbf{u}_a \in \mathbb{R}^3,
\label{eq:uav-kin}
\end{align}
where $\mathbf{u}_a$ directly commands the acceleration in each spatial dimension. This abstraction assumes that the UAV's attitude control system (inner-loop controller) can track desired accelerations sufficiently fast compared to the mission planning time scale.
Expanding the vectorized form, the dynamics becomes:
\begin{align}
\ddot{x}_a &= u_{a,x}, & 
\ddot{y}_a &= u_{a,y}, & 
\ddot{z}_a &= u_{a,z},
\label{eq:uav-expanded}
\end{align}
where
$
\mathbf{u}_a = [u_{a,x}, u_{a,y}, u_{a,z}]^\top
$
represents the accelerations commanded in the north-east-down coordinate frame. This model enables straightforward trajectory planning and control design while abstracting away the complex rotational dynamics and thrust vectoring mechanisms of the physical quadrotor. The yaw dynamics $\dot{\psi}_a$ is controlled independently through a separate yaw rate command, allowing the UAV to maintain the desired heading while following the planned trajectory.

\subsection{Mission Formulation and Planning with LLM}

The maritime inspection mission is formulated as a tuple:
\[
\centering
\langle \mathcal{M}, \mathcal{P}_{env},  \mathcal{K}, \mathcal{A}\rangle
\]

where, $\mathcal{M}$ contains the mission description (e.g., \textit{“Inspect the ship loading area, there should not be any human or vehicle near the ship loading crane, also make sure the container stacks are well aligned”}),
$\mathcal{P}_{env}$ describes the environmental state including the layout of the ship loading zone, workspace boundaries, crane position, cargo stacks, and dynamic elements.
$\mathcal{K}$ includes organized operational knowledge that covers safety and compliance guidelines, identified dangers in the loading zone, and navigation restrictions such as prohibited fly zones, maximum altitude limits, and areas where USVs are not permitted.
$\mathcal{A}$ is the action space, contains all possible actions that UAV and USV can take:

\[
\begin{aligned}
\mathcal{A} = \big\{
&\textit{Takeoff}, \ \textit{FlyTo}, \ \textit{Survey}, \ \textit{Record}, \ \textit{Hover}, \ \textit{Navigate}, \\
&\textit{Dock}, \ \textit{LandOnUSV}, \ \textit{Inspect}, \ \textit{Report}, \ \textit{GoHome}
\big\}
\end{aligned}
\]

A Large Language Model $\mathcal{L}$ establishes a mapping function that transforms the mission description into a symbolic mission plan $\pi$, mathematically represented as:
\begin{equation}
\label{eq:llm}
\mathcal{L}: \mathcal{M} \times \mathcal{P}_{env}
\times \mathcal{K}  \mapsto \pi
\end{equation}

The resulting symbolic plan: \( \pi = \langle \tau_1, \tau_2, \ldots, \tau_N \rangle \)
is constructed over the action space $\mathcal{A}$.

Each symbolic action is defined as:
\begin{equation}
\tau_i = \{a_i(\mathcal{R},\{ \theta_i\}, \{\sigma_i\}),p_i\}
\end{equation}
where, $a_i \in \mathcal{A}$ is the type of action, $\theta_i$ contains geometric and temporal parameters (target poses, survey patterns, dwell times), $\sigma_i$ specifies the sensor configurations and inspection criteria.
In general terms, \(\theta\) encodes the specification of \emph{``where and how long''}, by defining geometric targets and associated temporal durations. In contrast, \(\sigma\) defines \emph{``what to detect or inspect''}, by specifying the operational modes and the detection criteria
relevant to the inspection task, while $p_i$ denotes the set of preconditions necessary to meet before performing the $i$th action.

\subsection{VLM-based Inspection Overview}
The inspection system incorporates a vision language model $\mathcal{V}$ for intelligent scene understanding and compliance assessment. The VLM processes visual observations $\mathbf{I} \in \mathbb{R}^{H \times W \times C}$ (where $H$, $W$, and $C$ represent height, width, and color channels) along with contextual mission descriptions, such as, \textit{“Is there any human, vehicle (trucks, forklifts), or other suspicious objects near crane?"}, which provides context about what to look for. The VLM combines visual input with mission context, enabling it to focus on the ship loading area, recognize the crane, and detect the presence of relevant entities such as vehicles (e.g., trucks, forklifts) or humans operating nearby. Rather than returning only raw detections, the VLM produces a structured inspection report that specifies what objects or personnel were observed, their spatial location relative to the crane (e.g.; left side of the crane). This ensures that the output is directly interpretable and actionable for mission control.

The VLM mapping function is defined as:
\begin{equation}
\mathcal{V}: \mathbf{I} \times \mathcal{M}_{inspect}  \mapsto \mathcal{R}_{inspection}
\label{eq:vlmmod}
\end{equation}
where, $\mathbf{I}$ represents the captured visual data (images/video frames), $\mathcal{M}_{inspect}$ contains specific inspection objectives and a set of queries retrieved from $\sigma$ (which describe what to detect / inspect), $\mathcal{R}_{inspection}$ is the inspection report containing compliance status, detected anomalies and recommendations.

\subsection{Problem Statement}
Given a maritime port environment $\mathcal{P}_{env}$ with static and dynamic obstacles, a heterogeneous robotic system consisting of a USV and a UAV with complementary sensing capabilities, and a high-level inspection mission $\mathcal{M}$:

Develop an integrated autonomous inspection framework that:
\begin{itemize}
    \item Utilizes large language models to generate feasible symbolic plans $\pi$ that satisfy multi-platform constraints while achieving inspection objectives,
    \item Employs vision-language models to perform real-time scene understanding, compliance assessment, and automated report generation,
    \item Ensures safe and efficient multi-platform navigation through port environments while maintaining communication connectivity,
\end{itemize}

The system should maximize inspection coverage, ensure the accuracy of regulatory compliance assessment, minimize mission completion time, and maintain operational safety throughout the duration of the mission.

\begin{figure}[t]
    \centering
    \includegraphics[width=\linewidth]{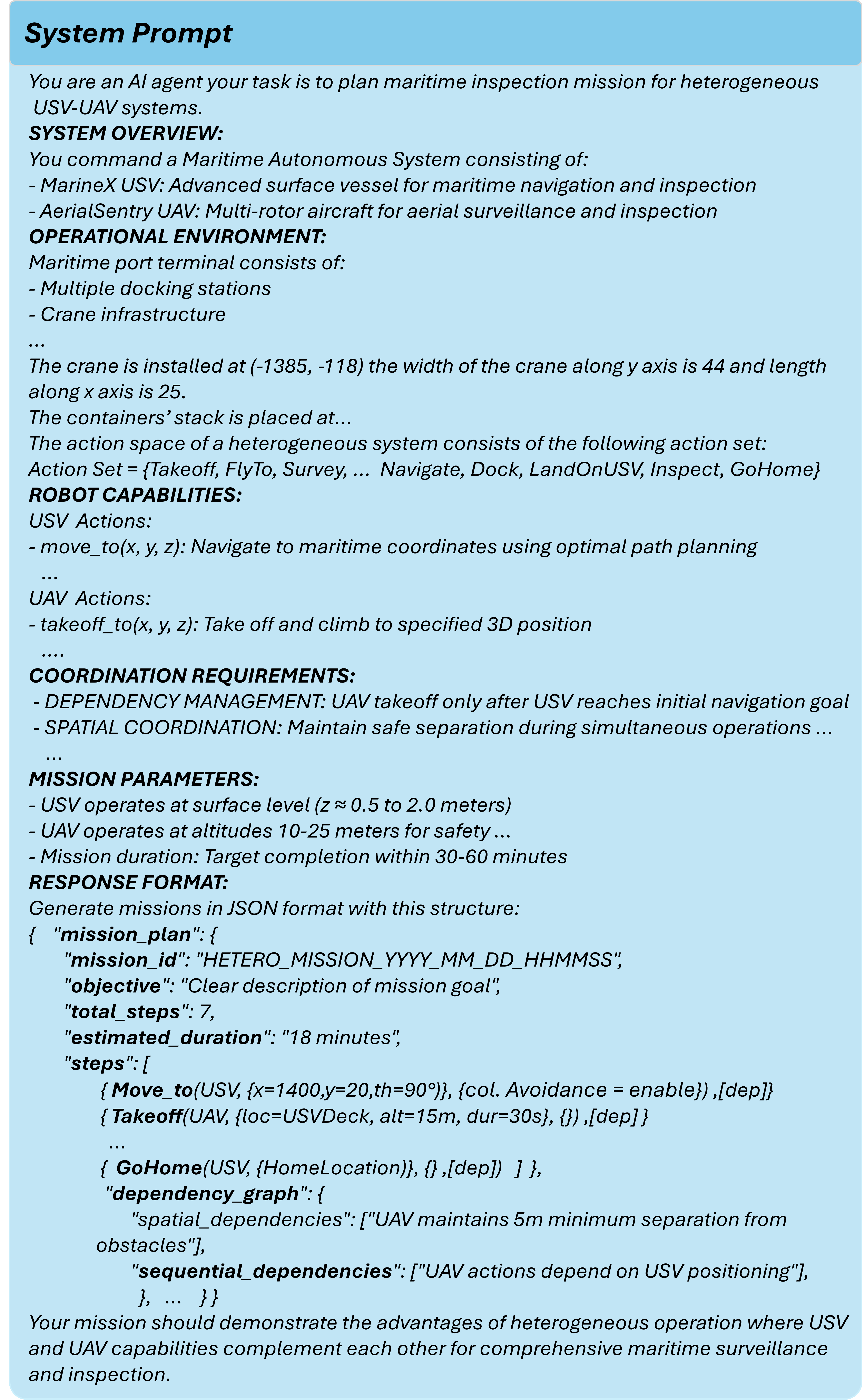}
    \caption{Example of a system prompt used for planning port inspection missions with heterogeneous USV-UAV system. The prompt defines the system overview, operational environment, robot capabilities, coordination requirements, mission parameters, and expected response format. It guides the LLM to generate symbolic mission plans with spatial and sequential dependencies, enabling coordinated heterogeneous operations for maritime inspection tasks.}
    \label{fig:prompt}
\end{figure}

\begin{figure}[t]
    \centering
    \includegraphics[width=\linewidth]{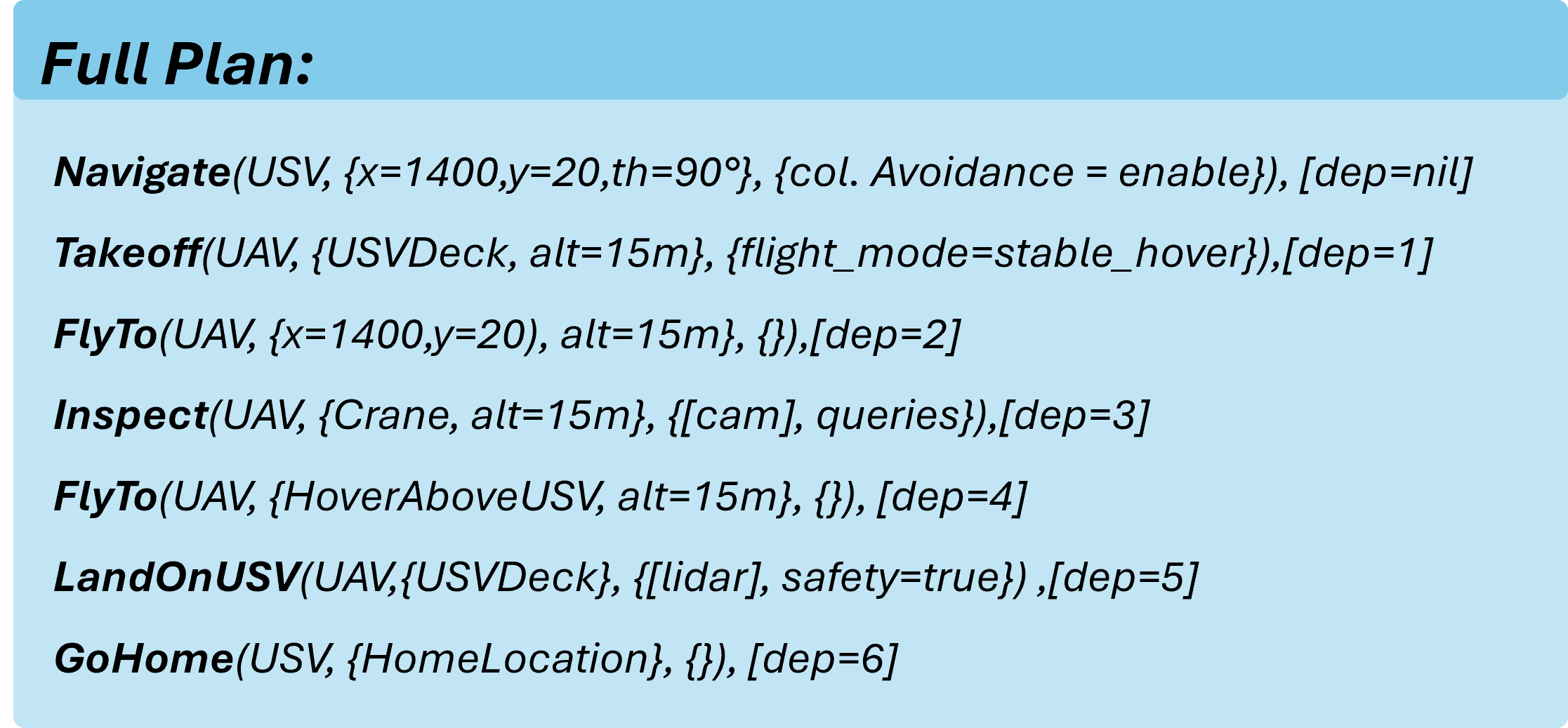}
    \caption{Example of a symbolic mission plan for a heterogeneous USV-UAV system generated by the LLM. The plan specifies sequential actions with explicit dependencies between tasks to ensure safe and coordinated execution of maritime inspection operations.}
    \label{fig:plan}
\end{figure}

\section{LLM-VLM Framework for Inspection}
\label{sec:framework}
In this section we will explain how the proposed LLM-VLM fusion framework integrates LLM for intelligent mission planning with VLMs for enhanced inspection in a heterogeneous UAV-USV maritime system. The architecture (Fig.~\ref{fig:framework}) comprises five components: (i) \emph{LLM Mission Planner}, (ii) a \emph{Communication Manager}, (iii) \emph{Heterogeneous robotic platforms} (\emph{USV} and \emph{UAV} systems), (iv) a \emph{VLM Inspector}, and (v) a centralized \emph{Control Center}.

\subsection{LLM Mission Planner Module}
The LLM Mission Planner serves as the high-level symbolic plan generator, using a pretrained LLM (e.g., GPT-4) as a black-box reasoning engine. It translates inspection tasks into executable symbolic plans for the heterogeneous system. These tasks are formulated by integrating mission instructions with the system prompt to create the final input provided to the LLM.

Mission instructions consist of text descriptions of a specific inspection tasks, such as:
\textit{Inspect the ship loading area; ensure no human or vehicle is near the loading crane, and verify that container stacks are well aligned.}
No special syntax is required in the mission instructions part of the prompt.
The system prompt contains a structured knowledge base that encodes operational restrictions, known hazards, regulatory rules (maritime safety and port regulations), platform capabilities/limits, and navigation restrictions. Prompt conditioning ensures that plans remain safe and feasible in maritime settings.
Text description of the environment and status summaries, workspace limits, static / dynamic obstacles (e.g., cranes, buildings, moving vessels), and current UAV/USV system states.
Fig.~\ref{fig:prompt} shows an example of some key parts of the system prompt.

The LLM processes these inputs through the mapping $\mathcal{L}(\cdot)$ defined earlier (Eq.~\ref{eq:llm}) and generates a symbolic plan $\pi$ drawn from the action space $\mathcal{A}$. The plan includes coordinated actions across platforms. Fig.~\ref{fig:plan} shows an example symbolic mission plan generated by the LLM for crane area inspection.  

In this mapping, the geometric and temporal parameters $\theta$ (e.g., target location, altitude, duration) guide the formation of waypoints and constraints, while the sensing and inspection descriptors $\sigma$ contain a set of predefined queries for the VLM (e.g., \textit{``Is there any vehicle near the crane?''}, \textit{``Is there any human in the crane loading area?''}) that are used to populate the inspection schedule. Additionally, the symbolic plan also define the preconditions for each action in the form of dependancy graph, as detailed in Sec.~\ref{com-man}.

\subsection{Communication Manager}\label{com-man}

The communication manager is responsible for UAV-USV heterogeneous system coordination using dependency graph to enable safe and efficient execution of complex multi-robot operations through dynamic dependency resolution and parallel task execution.

The dependency graph in our heterogeneous USV-UAV mission coordination system represents a directed acyclic graph  where nodes correspond to individual symbolic actions
\( a_i(\mathcal{R}, \{\theta_i\}, \{\sigma_i\})\)
and edges encode temporal and spatial constraints between mission steps. The graph construction algorithm analyzes the generated mission plan and automatically identifies two primary types of dependencies (preconditions):
(1) \emph{Inter-robot spatial preconditions}, where UAV actions such as \(\textsc{Takeoff}(\mathrm{UAV}, \{\mathrm{USVDeck},\ \mathrm{alt}=15\text{ m}\}, \{\})\) require the USV to be positioned at a specific location through a preceding \(\textsc{Navigate}(\mathrm{USV}, \{\mathrm{position}\}, \{\})\) action;
(2) \emph{Sequential robot-specific preconditions}, ensuring that each robot completes its current action before proceeding to the next, maintaining state consistency and preventing command conflicts.
The dependency resolution mechanism employs a topological sorting approach combined with real-time status monitoring, where each completed action \(\tau_i\) triggers a re-evaluation of all pending actions to identify newly executable steps whose precondition prerequisites have been satisfied.

\begin{table}[h!]
\centering
\renewcommand{\arraystretch}{1.15}
\setlength{\tabcolsep}{4pt} 
\scriptsize
\caption{Execution timeline derived from the dependency graph for heterogeneous USV--UAV mission coordination. Each action \(\tau_i\) is executed only after its preconditions are satisfied, ensuring correct sequencing and safe parallel operations.}
\label{tab:execution-timeline}
\begin{tabular}{lllp{2.5cm}} 
\toprule
\textbf{Time} & \textbf{Action} & \textbf{Robot} & \textbf{Preconditions Status} \\
\midrule
\(t_0\) & Execute \(\tau_0\) & USV Navigate & No preconditions \\
\(t_1\) & Complete \(\tau_0\) \(\rightarrow\) Execute \(\tau_1\) & UAV Takeoff & \(\mathrm{precond}(\tau_1)=\{\tau_0\}\ \checkmark\) \\
\(t_2\) & Complete \(\tau_1\) \(\rightarrow\) Execute \(\tau_2\) & UAV FlyTo & \(\mathrm{precond}(\tau_2)=\{\tau_1\}\ \checkmark\) \\
\(t_3\) & Complete \(\tau_2\) \(\rightarrow\) Execute \(\tau_3\) & UAV Inspect & \(\mathrm{precond}(\tau_3)=\{\tau_2\}\ \checkmark\) \\
\(t_4\) & Complete \(\tau_3\) \(\rightarrow\) Execute \(\tau_4\) & UAV FlyTo (Return) & \(\mathrm{precond}(\tau_4)=\{\tau_3\}\ \checkmark\) \\
\(t_5\) & Complete \(\tau_4\) \(\rightarrow\) Execute \(\tau_5\) & UAV LandOnUSV & \(\mathrm{precond}(\tau_5)=\{\tau_4\}\ \checkmark\) \\
\(t_6\) & Complete \(\tau_5\) \(\rightarrow\) Execute \(\tau_6\) & USV GoHome & \(\mathrm{precond}(\tau_6)=\{\tau_5\}\ \checkmark\) \\
\(t_7\) & Complete \(\tau_6\) & Mission End & All preconditions satisfied \\
\bottomrule
\end{tabular}
\end{table}

The mission coordinator implements a dynamic execution engine that continuously monitors the dependency graph state and executes all ready actions in parallel when their preconditions are satisfied, maximizing system efficiency while maintaining safety constraints. The coordination logic maintains a completion tracking system using a set-based approach where
\(\mathrm{completed\_steps} = \{\, i \mid \tau_i \in \mathrm{executed\_actions} \,\}\),
and for each pending action \(\tau_j\), the system verifies that all preconditions \(p_j \subseteq \mathrm{completed\_steps}\) before initiating execution. This approach enables sophisticated coordination patterns, such as allowing the UAV to perform \(\textsc{Survey}(\mathrm{UAV}, \{\mathrm{Orbit360},\ \mathrm{alt}=15\text{ m}\}, \{\mathrm{cam}\})\) while allowing the USV to execute $\textsc{Navigate}(\mathrm{USV}, \{\mathrm{next\_position}\}, \{\})$ to the next waypoint, provided there are no spatial conflicts. The dependency graph also handles error recovery and mission re-planning scenarios, where failed actions can trigger precondition re-evaluation and alternative execution paths. Critical synchronization points, such as the transition from autonomous UAV operations back to USV-deck landing, are managed through explicit precondition chains that ensure proper sequencing of
\begin{flalign*}
\textsc{FlyTo}(\mathrm{UAV}, \{\mathrm{HoverPointAboveUSV}\}, \{\}) 
&\;\rightarrow\; &\\
\textsc{LandOnUSV}(\mathrm{UAV}, \{\mathrm{USVDeck}\}, \{\mathrm{safety\_checks}\}) 
&\;\rightarrow\; &\\
\textsc{GoHome}(\mathrm{USV}, \{\mathrm{PortDock}\}, \{\}) \,, &&
\end{flalign*}

ensure mission safety and operational reliability throughout the heterogeneous system coordination process.

The execution timeline presented in Table~\ref{tab:execution-timeline} illustrates how the dependency graph enforces the correct sequence of actions while allowing efficient parallelization when possible. Each row corresponds to a symbolic action \(\tau_i\), with the column \emp{Preconditions Status} showing that an action is only triggered once all its dependencies are satisfied. This structured execution log makes explicit the topological ordering derived from the dependency graph and demonstrates how the UAV and USV coordinate seamlessly across navigation, takeoff, survey, reporting, return, landing, and docking phases. By aligning symbolic actions with their temporal order and dependency checks, the table validates the safety-critical synchronization ensured by the proposed coordination framework.

\begin{figure}[t]
    \centering
    \includegraphics[width=0.85\linewidth]{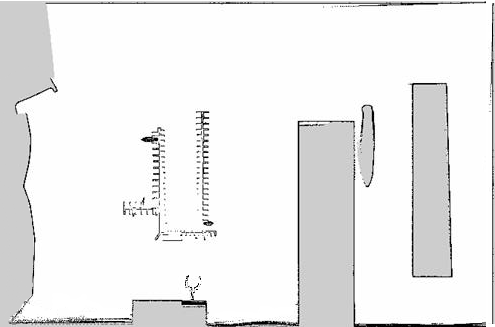}
    \caption{Map of the simulated port environment used for USV navigation and inspection tasks. The environment includes multiple docking stations, crane areas, container stacks, and open water regions, providing a realistic maritime setting to evaluate autonomous surface vehicle path planning and mission execution.}
    \label{fig:slam-map}
\end{figure}

\subsection{Heterogeneous Platform Integration}
In this section, we will describe the UAV-USV heterogeneous system architectures, sensing, and control of the platforms for maritime inspection. 
\subsubsection{USV System Architecture}
The USV system serves as the primary coordination platform and mobile operational base for the heterogeneous maritime inspection framework, providing the stability necessary for extended autonomous missions in challenging port environments. The USV perception module integrates multiple sensor modalities, including a camera for inspection, inertial measurement units (IMU) for heading correction, and LiDAR sensors for obstacle detection, localization, and mapping. The LiDAR provides 270 degree scanning that enables robust detection of both static infrastructure and dynamic obstacles, such as moving vessels, while simultaneously supporting simultaneous localization and mapping (SLAM) capabilities essential for autonomous navigation in complex port environments.

The navigation subsystem uses the NAV2 framework, which provides a comprehensive solution for autonomous navigation. The system constructs and maintains real-time maps of the port environment using LiDAR data, enabling precise localization; Fig.~\ref{fig:slam-map} shows the generated map of the example port scene. The SLAM algorithm continuously updates the occupancy grid map that represents the probability of the presence of obstacles in each grid cell. This real-time update mechanism ensures that the USV maintains situational awareness even under dynamic maritime conditions.

\begin{figure}[t]
    \centering
    \includegraphics[width=0.9\linewidth]{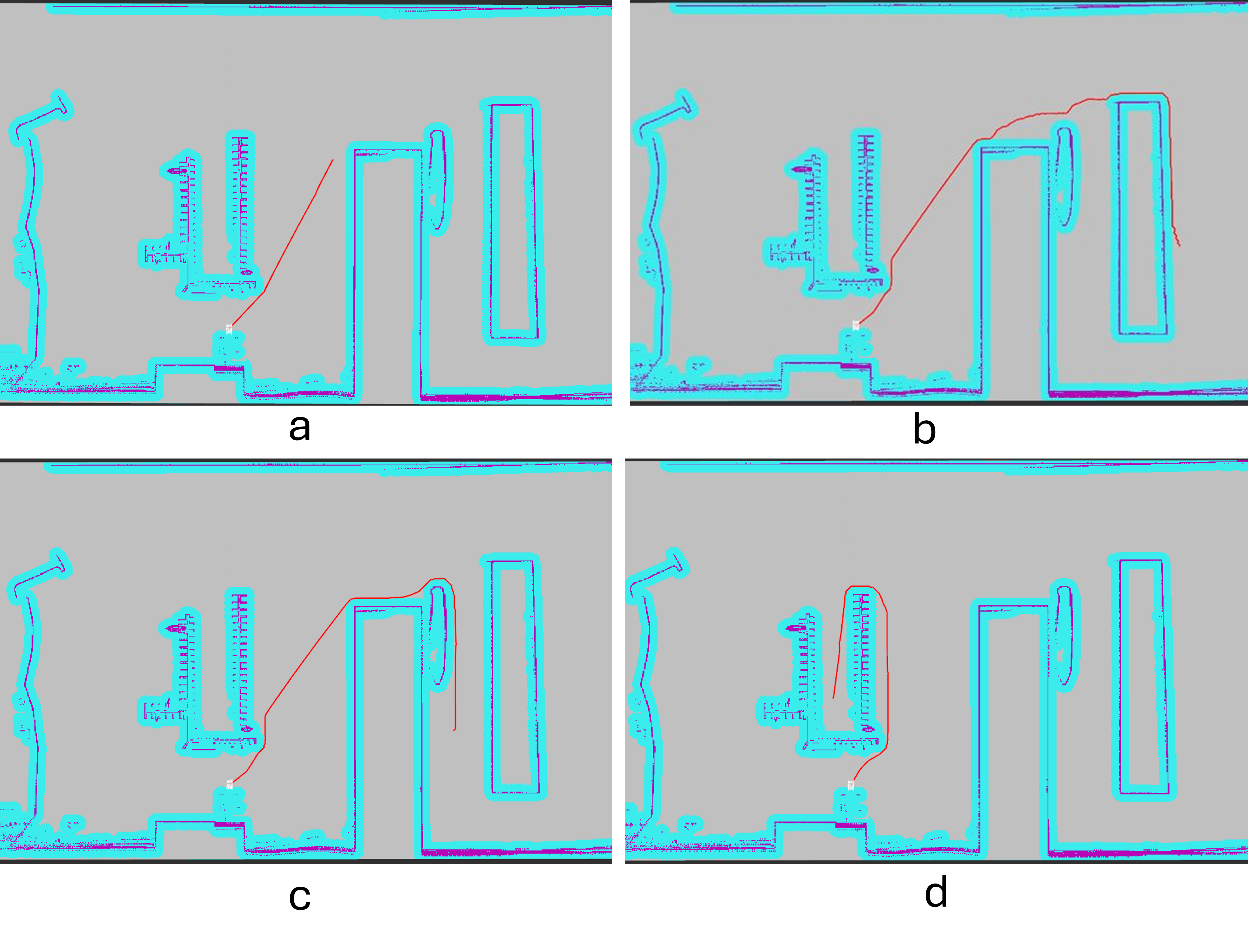}
    \caption{Example USV navigation paths in the simulated port environment. The red lines indicate the planned trajectories for different inspection missions, including navigation to the docking area (top left), traversal toward the container stack (top right), movement around the crane infrastructure (bottom left), and inspection of docking stations (bottom right). The background map highlights the port layout, providing realistic constraints for autonomous path planning and mission execution.}
    \label{fig:paths-a}
\end{figure}

The navigation system employs the A* path planning algorithm for global route optimization, generating optimal paths from the current position to the target while considering obstacle constraints; Fig.~\ref{fig:paths-a} shows some sample path generated for USV navigation. The A* planning strategy balances the travel distance with the safety margins by penalizing paths that pass too close to static or dynamic obstacles, ensuring both efficiency and robustness in mission execution.

The USV control system employs a PID-based feedback controller to precisely track trajectory, compensating for environmental disturbances. The controller regulates both linear and angular velocities to ensure accurate path following and supports precise station-keeping capabilities, which are essential for UAV launch and recovery operations. By maintaining position accuracy within bounds, the control system enables safe heterogeneous coordination between UAV and USV. Finally, the USV actuator system provides the differential thrust mechanisms that enable precise maneuvering in confined port environments, executing the velocity commands generated by the navigation layer while supporting the complex positioning requirements necessary for heterogeneous platform coordination.
\begin{figure}[t]
    \centering
    \includegraphics[width=\linewidth]{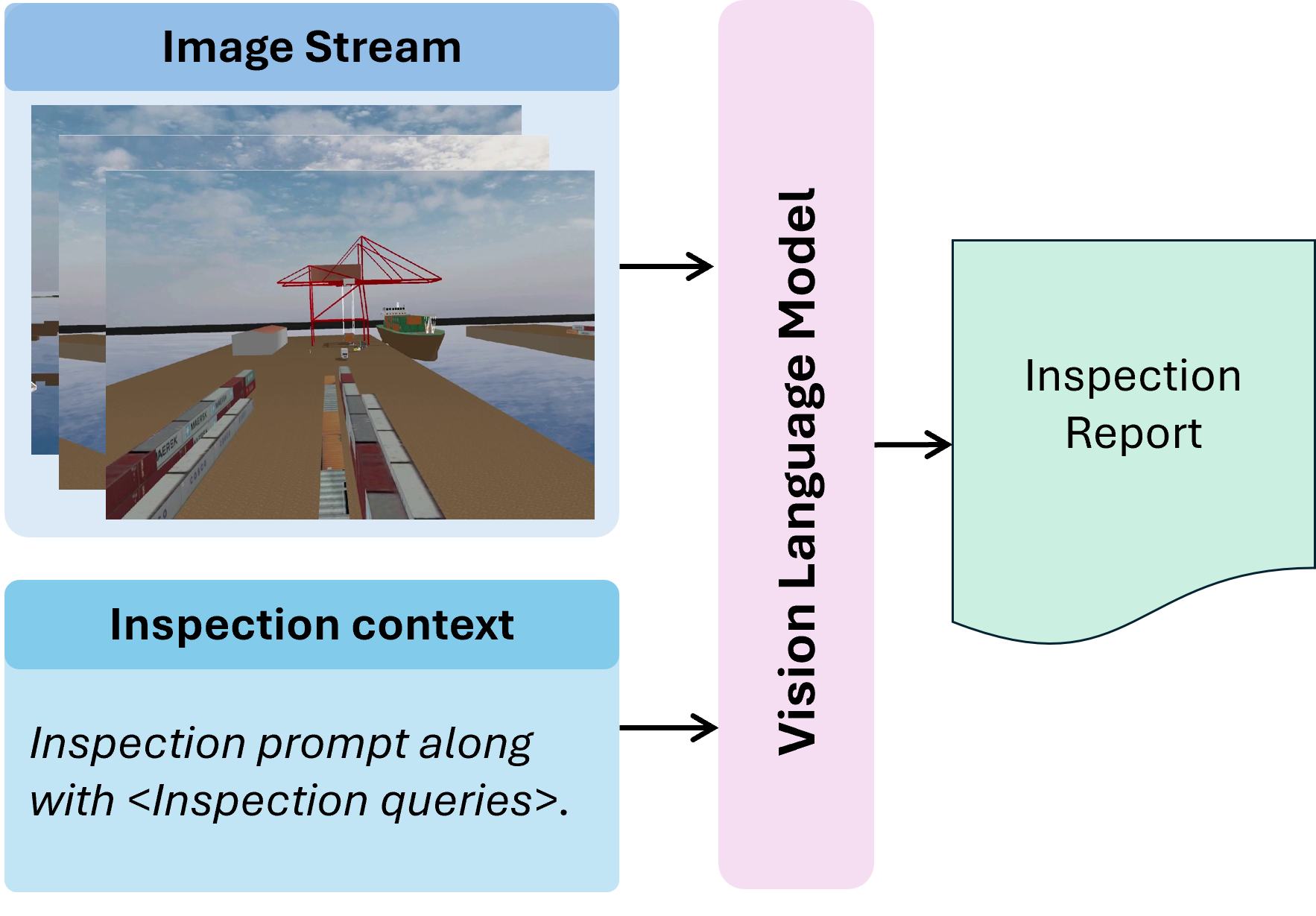}
    \caption{VLM-based inspection pipeline. The system processes an image stream of the port environment along with the inspection context containing predefined queries. The VLM interprets the combined input and generates a structured inspection report highlighting relevant findings for maritime safety and monitoring.}
    \label{fig:paths-a}
\end{figure}

\subsubsection{UAV System Architecture}
The UAV system provides essential aerial surveillance capabilities and access to elevated inspection points that are inaccessible to surface vessels, extending the operational range of the heterogeneous inspection system. The UAV perception module incorporates a high-resolution camera system that provides imagery for visual inspection of port infrastructure from optimal viewpoints. An onboard IMU supplies orientation data, while the UAV’s position is derived from the state in the simulator (Gazebo). This information is used both for accurate position estimation and for enabling precise landing operations.  

The UAV waypoint and line-of-sight guidance system manages three-dimensional trajectory following with optimized altitude and heading control for maritime inspection missions. It applies position-based guidance for waypoint navigation, enabling the UAV to follow reference paths accurately while compensating for environmental disturbances. For landing operations, the USV is assumed to be stationary at a known location during the UAV’s approach and touchdown. Under this assumption, the UAV follows fixed descent trajectories aligned with the deck coordinates of the USV, simplifying landing execution.  

The UAV control architecture employs cascaded PID controllers for attitude and position regulation. Inner-loop attitude control provides rapid stabilization, while outer-loop position control ensures smooth trajectory execution under maritime wind conditions. The control system regulates roll, pitch, and yaw angles to maintain stable flight during inspection tasks, while the actuator system provides thrust and control surface adjustments that allow precise hovering and maneuvering. This architecture ensures that the UAV can maintain stability during inspections and perform complex maneuvers such as coordinated landings on the stationary USV platform.  
\begin{figure}[t]
    \centering
    \includegraphics[width=\linewidth]{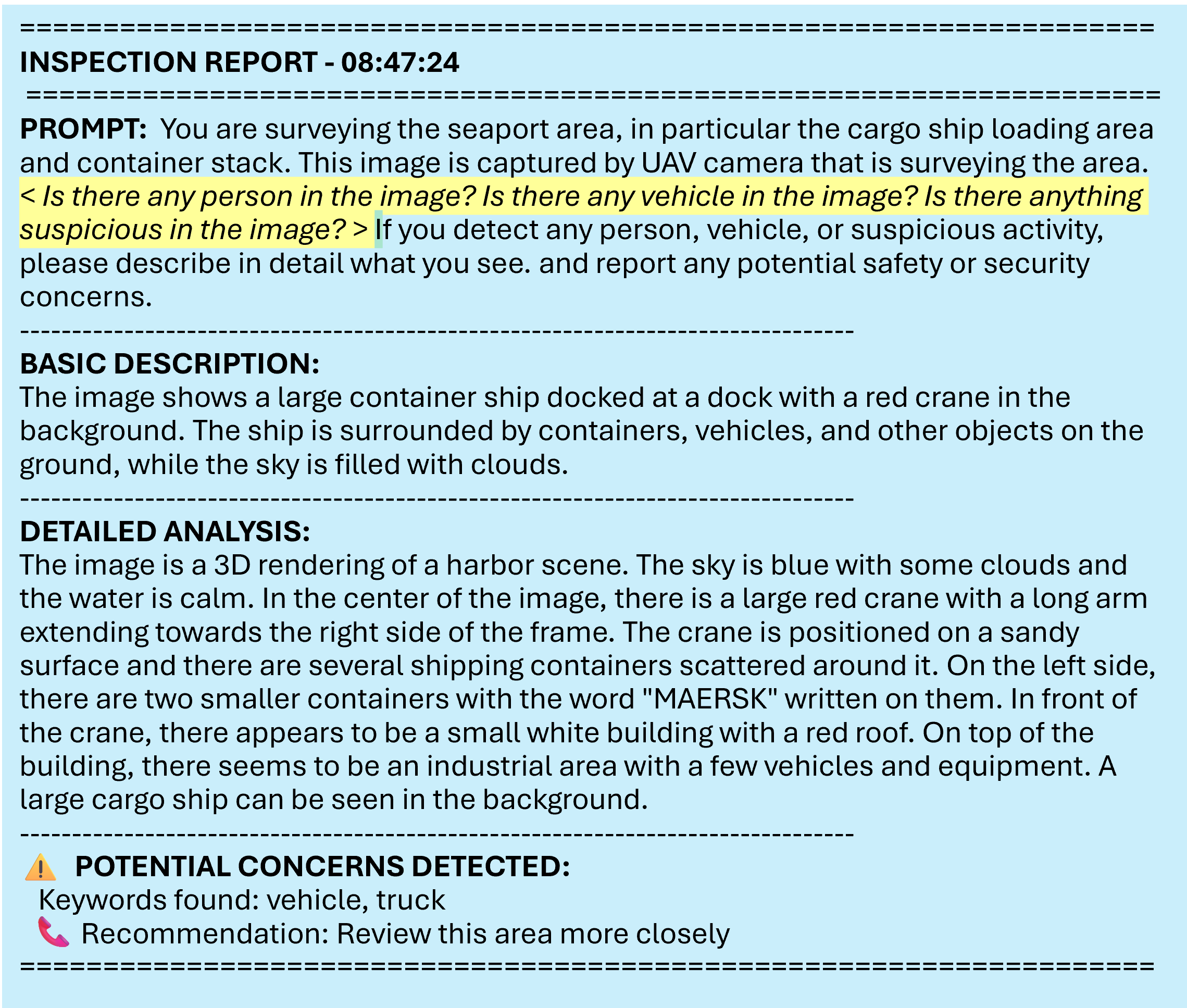}
    \caption{Example surveillance report generated by the VLM during a port inspection mission. The UAV-captured image stream is analyzed against predefined queries (e.g., presence of people, vehicles, or suspicious activity). The output includes a basic description, detailed analysis, and potential concerns, providing structured situational awareness and actionable recommendations for maritime security monitoring.}
    \label{fig:insp-report}
\end{figure}
\subsection{VLM-based Inspection}
The VLM-driven inspection system has a significant advantage over traditional computer vision approaches by incorporating semantic understanding and contextual reasoning capabilities. 
As explained in Eq.~\ref{eq:vlmmod}, the module processes visual observations $\mathbf{I}\!\in\!\mathbb{R}^{H\times W\times C}$ from the USV and UAV platforms independently. Unlike classical computer vision methods that recognize a fixed set of visual patterns, the VLM Inspector fuses multi-modal visual and textual input by integrating images with mission descriptions and regulatory knowledge. For example, when instructed to \emph{“inspect the ship loading area; there should not be any human or vehicle near the ship-loading crane”} $\mathcal{V}$ applies the mapping function in Eq.~\ref{eq:vlmmod} to produce structured findings.

To operate reliably, the VLM $\mathcal{V}$ requires a mission-specific natural language prompt that grounds the scene, site rules, and reporting requirements. For the above port inspection mission, the VLM inspection prompt $\mathcal{M}_\text{Inspect}$ used is shown in the first part of the Fig.~\ref{fig:insp-report}, the part highlighted in yellow corresponds to the queries obtained from $\sigma$.

Once a mission specific prompt is provided, the VLM processing pipeline proceeds in three stages: (i)~\textit{visual and text encoding}, (ii)~\textit{multimodal fusion}, and (iii)~\textit{text generation}. In the encoding stage, raw image data are transformed into visual feature embeddings via a visual encoder, while the text encoder processes inspection instructions to form textual embeddings. These modalities are aligned during multimodal fusion through a cross-attention mechanism, allowing visual features to focus on the textual context. This ensures that perception is guided by inspection goals. Finally, the fused representation is passed to a language model decoder that produces a structured natural language inspection report, detailing observed anomalies and compliance assessments grounded in both the visual evidence and mission specific constraints.

The primary benefit of VLMs is their ability to interpret maritime scenes with a semantic understanding that mimics human reasoning in maritime contexts. Beyond object detection, the VLM Inspector generates structured inspection reports
$\mathcal{R}_{\text{inspection}}$ that follows the formate as shown below.

\begin{equation}
\mathcal{R}_{\text{inspection}} = 
\{\mathcal{P}_{\text{prompt}}, 
\mathcal{D}_{\text{basic}}, 
\mathcal{A}_{\text{detailed}}, 
\mathcal{C}_{\text{concerns}}\}
\end{equation}

where $\mathcal{P}_{\text{prompt}}$ is the mission specific instruction used to contextualize inspection,  $\mathcal{D}_{\text{basic}}$ provides a short scene-level summary,  $\mathcal{A}_{\text{detailed}}$ provides a comprehensive analysis of observed infrastructure, vessels, and activities, and  $\mathcal{C}_{\text{concerns}}$ lists anomalies, safety issues, or regulatory violations with suggested actions.  
Fig.~\ref{fig:insp-report} show the inspection report generated for an inspection mission. B

\subsection{Control Center}
The Control Center acts as the central supervisory hub for the entire maritime inspection framework, allowing real-time mission tracking and dynamic coordination between USV and UAV platforms. Continuously monitors the health of each platform, the progress of assigned inspection tasks, and the outputs generated by the VLM. This live monitoring ensures that operations remain safe and mission objectives are achieved effectively. When unexpected environmental changes or anomalies are detected, such as; adverse weather, or inspection critical findings, the Control Center initiates adaptive re-planning procedures by triggering the LLM Mission Planner to generate updated symbolic mission plans.

In addition to supervision and re-planning, the Control Center is responsible for aggregating inspection results into comprehensive, structured reports. These reports combine VLM generated findings with contextual metadata such as timestamps and platform specific data, ensuring traceability and regulatory compliance. 
\begin{algorithm}
\caption{LLM-VLM Fusion Framework Execution Flow}\label{algo}
\begin{algorithmic}[1]
\REQUIRE $\mathcal{M}_{\text{instruction}}, \mathcal{P}_{\text{env}}, \mathcal{K}, \mathcal{A}$
\ENSURE $\mathcal{R}_{\text{Inspection}}, \text{Mission\_status}$
\STATE Initialize USV/UAV and communication links
\STATE $\pi \leftarrow \mathcal{L}(\mathcal{M}_{\text{instruction}}, \mathcal{P}_{\text{env}}, \mathcal{C}_{\text{constraints}}, \mathcal{K}_{\text{knowledge}})$
\STATE {$G \leftarrow$ \text{BuildDependencyGraph} \;{$\pi$}};  
\STATE $\mathit{completed},\mathit{executing},\mathit{inspection\_results} \leftarrow \emptyset$
\WHILE{$|\mathit{completed}| < |\pi|$}
    \STATE $\mathit{ready} \leftarrow$ actions in $\pi$ whose preconditions are satisfied
    \FORALL{$\tau_j \in \mathit{ready}$}
        \STATE Execute UAV\/USV motion; \text{ExecuteParallel}{$\tau_j$}; update $\mathit{executing}$
        \IF{$\tau_j$ is inspection}
            \STATE \hspace{0pt}\makebox[0pt][l]{$\mathbf{I} \leftarrow$ cam stream;}
            \STATE $\mathcal{R}_{\text{inspection}} \leftarrow \mathcal{V}(\mathbf{I}, \mathcal{M}_{\text{inspect}})$
            \IF{critical\_issue($\mathcal{R}_{\text{inspection}}$)} \STATE alert\_operators() \ENDIF
            \STATE Append $\mathcal{R}_{\text{inspection}}$ to $\mathit{inspection\_results}$
        \ENDIF
    \ENDFOR
    \STATE Update $\mathit{completed}, \mathit{executing}, G$ from finished tasks
    \STATE $\mathcal{P}_{\text{env}} \leftarrow$ update environment; 
           \IF{obstacles or degraded comms}
           \STATE replan $\pi$, rebuild $G$ 
           \ENDIF
\ENDWHILE
\STATE Execute safe-return; $\mathcal{R}_{\text{final}} \leftarrow \mathit{inspection\_results}$
\RETURN $\mathcal{R}_{\text{final}}, \text{Mission\_status}$
\end{algorithmic}
\end{algorithm}

\begin{figure*}[t]
    \centering
    \includegraphics[width=\linewidth]{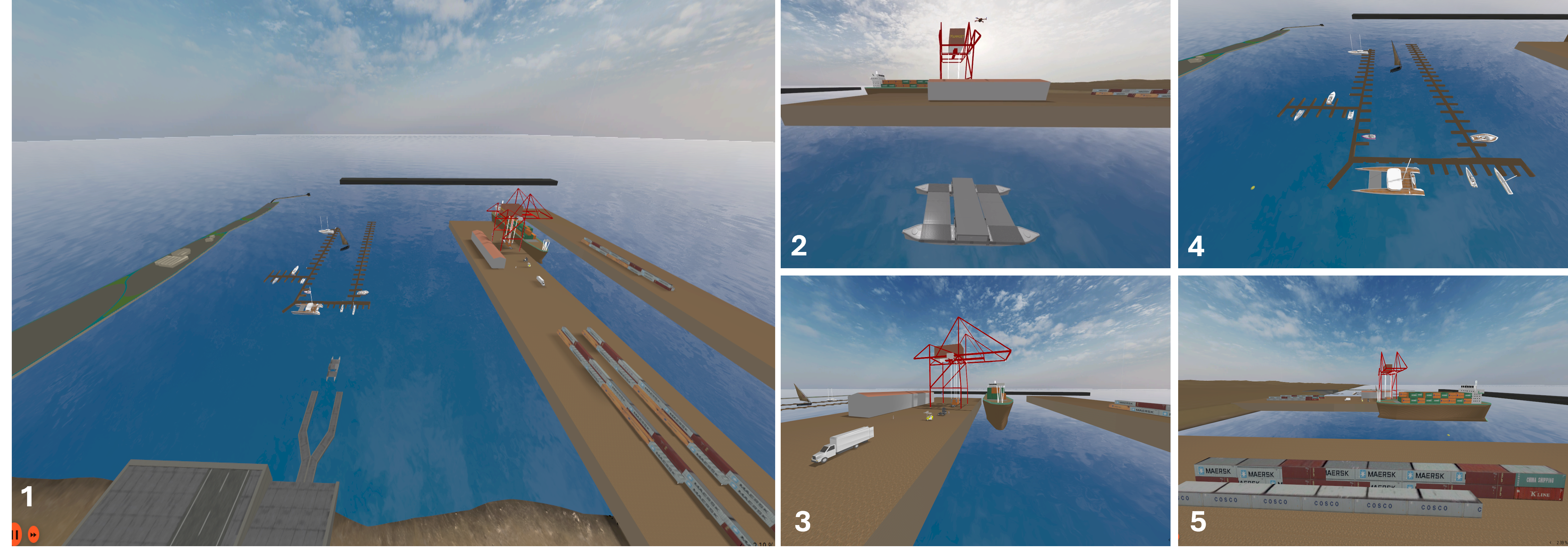}
    \caption{Snapshots of the simulation environment for heterogeneous port inspection missions. (1) Complete view of the maritime environment; (2) UAV conducting crane inspection while the USV remains at the UAV take-off location; (3) port crane with cargo ships and vehicles nearby; (4) central docking station with multiple boats; (5) container stacks positioned near the crane area.Model Loading time}
    \label{fig:snap}
\end{figure*}

\subsection{Framework Operation Flow Explanation}

The operational workflow of the proposed inspection framework is summarized in Algorithm~\ref{algo}. First, the LLM processes the mission instructions $\mathcal{M}_{\text{instruction}}$, environmental context $\mathcal{P}_{\text{env}}$, prior knowledge $\mathcal{K}_{\text{knowledge}}$, and $\mathcal{A}$ action space to produce a symbolic plan $\pi$. A dependency graph $G$ is then constructed, while the sets for completed tasks, executing tasks, and inspection results are initialized.

During execution, the framework selects actions whose preconditions are satisfied and sends them to the USV and UAV. If an action involves inspection, sensor data $\mathbf{I}$ are captured and processed by the VLM $\mathcal{V}$ using inspection objectives $\mathcal{M}_{\text{inspect}}$, resulting in an inspection report $\mathcal{R}_{\text{inspection}}$. Critical findings trigger immediate alerts, and all results accumulate.

The environment state $\mathcal{P}_{\text{env}}$ is continuously updated and in case failure of the subtask or other emergency conditions, the LLM replans the remaining tasks and rebuilds the dependency graph. Once all actions are completed, a safe return protocol is executed, and the final report $\mathcal{R}_{\text{final}}$, the collection of all inspection results is transmitted with the mission status to the control center.

\begin{table*}[h!]
\centering
\caption{Benchmark results of LLM models for symbolic mission plan generation across inspection tasks of increasing complexity. Performance is evaluated in terms of response time (RT, in seconds), correctness (Corr., 0–100), and execution success rate (Succ., in \%). The reported values are derived from experimental trials conducted over multiple heterogeneous inspection scenarios.}\label{table-llm}
\setlength{\tabcolsep}{5pt} 
\renewcommand{\arraystretch}{1.4} 
\resizebox{\textwidth}{!}{%
\begin{tabular}{|p{6.8cm}|
c|c|c||c|c|c||c|c|c||c|c|c||c|c|c|}
\hline
\multirow{2}{*}{\textbf{Task}} 
& \multicolumn{3}{c||}{\textbf{GPT-4o}} 
& \multicolumn{3}{c||}{\textbf{GPT-3.5-Turbo}} 
& \multicolumn{3}{c||}{\textbf{GPT-4}} 
& \multicolumn{3}{c||}{\textbf{Gemini}} 
& \multicolumn{3}{c|}{\textbf{LLaMA}} 
\\ \cline{2-16}
& \textbf{RT} & \textbf{Corr.} & \textbf{Succ.} 
& \textbf{RT} & \textbf{Corr.} & \textbf{Succ.} 
& \textbf{RT} & \textbf{Corr.} & \textbf{Succ.} 
& \textbf{RT} & \textbf{Corr.} & \textbf{Succ.} 
& \textbf{RT} & \textbf{Corr.} & \textbf{Succ.} 
\\ \hline\hline
Inspect the central docking zone and detect unauthorized sailboats (Task1) 
& 7.8 & 100 & 90.0 & 7.7 & 100 & 95.0 & 13.9 & 100 & 85.0 & 12.7 & 100 & 95.0 & 9.8 & 48 & 35.0 \\ \hline
Inspect unauthorized personnel/vehicles in the crane work zone (Task2) 
& 8.5 & 100 & 95.0 & 8.3 & 100 & 90.0 & 15.1 & 100 & 95.0 & 13.3 & 100 & 85.0 & 10.5 & 46 & 40.0 \\ \hline
Survey the container stack and check the stacking (Task3) 
& 7.2 & 100 & 95.0 & 8.8 & 100 & 90.0 & 13.0 & 100 & 90.0 & 14.0 & 100 & 96.7 & 11.4 & 44.8 & 41.9 \\ \hline
Inspect the docking + container stack in rectangular form (Task4) 
& 10.0 & 86.5 & 75.0 & 9.3 & 83.8 & 70.0 & 15.4 & 83.3 & 70.0 & 14.3 & 71.6 & 50.0 & 10.3 & 34.7 & 20.0 \\ \hline
Joint UAV-USV inspection, crane + docking station  (Task5) 
& 8.7 & 85.0 & 75.0 & 9.6 & 82.1 & 75.0 & 14.9 & 82.4 & 65.0 & 13.5 & 52.0 & 35.0 & 11.5 & 40.0 & 25.0 \\ \hline\hline
\textbf{Average across tasks} 
& 8.44 & 94.3 & 86.0 & 8.74 & 93.2 & 84.0 & 14.46 & 93.1 & 81.0 & 13.56 & 84.7 & 72.3 & 10.7 & 42.7 & 32.4 \\ \hline
\end{tabular}
}
\end{table*}
\section{Experimental Validation}

\subsection{Experimental Setup}
To evaluate the proposed approach, we conducted extensive testing using the MBZIRC Maritime Simulator, an open source ROS2-based platform developed in C++ and Python for the Mohamed Bin Zayed International Robotics Competition (MBZIRC). This simulator offers advanced capabilities for maritime robotics, allowing for the complete validation of autonomous navigation and inspection frameworks in realistic and challenging conditions.
The key strength of the simulator lies in its ability to model complex hydrodynamic and hydrostatic behaviors of surface vehicles while incorporating environmental dynamics such as waves, currents, and wind forces. 

We incorporate heterogeneous UAV-USV systems for autonomous port inspection, allowing joint aerial-surface cooperation. This integration allowed us to replicate complex maritime missions in which UAVs provided high-altitude situational awareness, rapid deployment, and overhead inspection, while USVs offered stable long-duration operation, close-range infrastructure inspection.
To further enhance realism, we developed a detailed virtual port infrastructure within the simulator. The environment includes a harbor basin, multiple types of boats, cargo ships, ship-loading cranes, ground vehicles (e.g., trucks and forklifts), and stacked shipping containers, effectively replicating an operational port, some of the snapshots of the simulation environment are shown in Fig.~\ref{fig:snap}. This setup allowed us to evaluate UAV-USV collaboration under representative conditions for port inspection, monitoring, and logistics operations.

Using this extended simulator, we tested the system in several heterogeneous mission tasks, demonstrating cooperative capabilities in port inspection and operational safety. Some of the examples of these mission tasks are summarized in Table~1, which highlights the diversity of UAV-USV coordination scenarios considered during evaluation.

\subsection{LLM-based Mission Planning Results and Analysis}  
\begin{figure}[t]
    \centering
    \includegraphics[width=\linewidth]{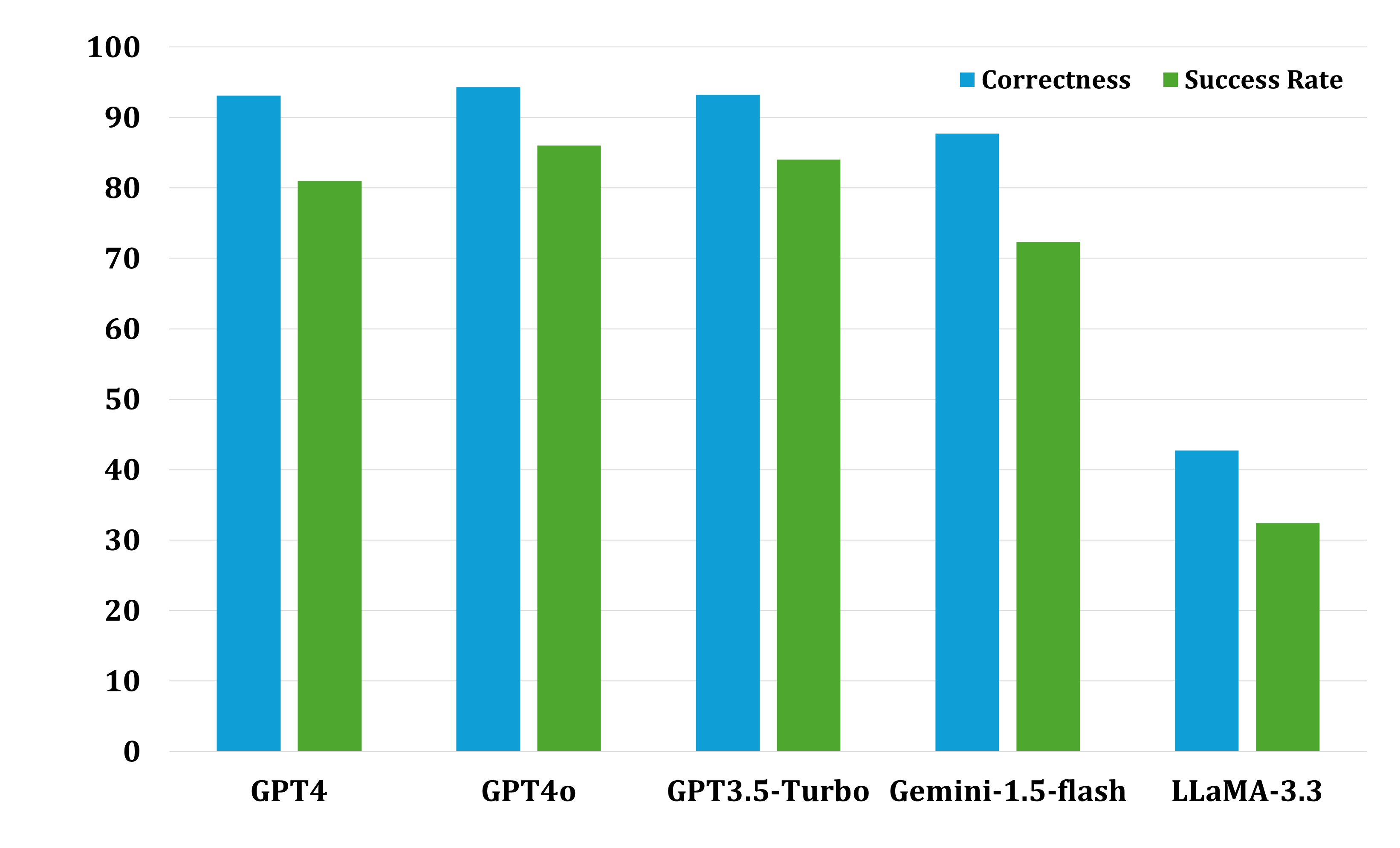}
    \caption{Benchmark comparison of LLM models for symbolic mission plan generation. Performance is described in terms of average correctness and execution success rate across all the performed tasks. GPT-4, GPT-4o, and GPT-3.5-Turbo achieve consistently high correctness and success rates, Gemini-1.5-flash shows moderate performance, while LLaMA-3.3 exhibits significantly lower values across both metrics.}
    \label{fig:llm-success}
\end{figure}

To evaluate different LLMs for maritime mission planning, we benchmarked GPT-4o,GPT-4, GPT-3.5-Turbo, Gemini and LLaMA. These models were tested in a set of heterogeneous UAV-USV inspection missions of varying complexity.  
We evaluated performance using three key indicators:  
\begin{itemize}  
    \item \textbf{Correctness}: each task was evaluated by two robotic experts and assigned a score (0-100) distributed across three criteria (i) valid JSON structure (20 points), (ii) mission tasks in correct order (40 points), and (iii) correct specification of task preconditions in the dependency graph (40 points).  
    \item \textbf{Execution Success Rate}: the percentage of times the generated plan was executed successfully by the heterogeneous UAV-USV system in simulation.  
    \item \textbf{Response Time (RT)}: the average API response time required for the model to generate a complete mission plan.  
\end{itemize}

\begin{figure}[t]
    \centering
    \includegraphics[width=\linewidth]{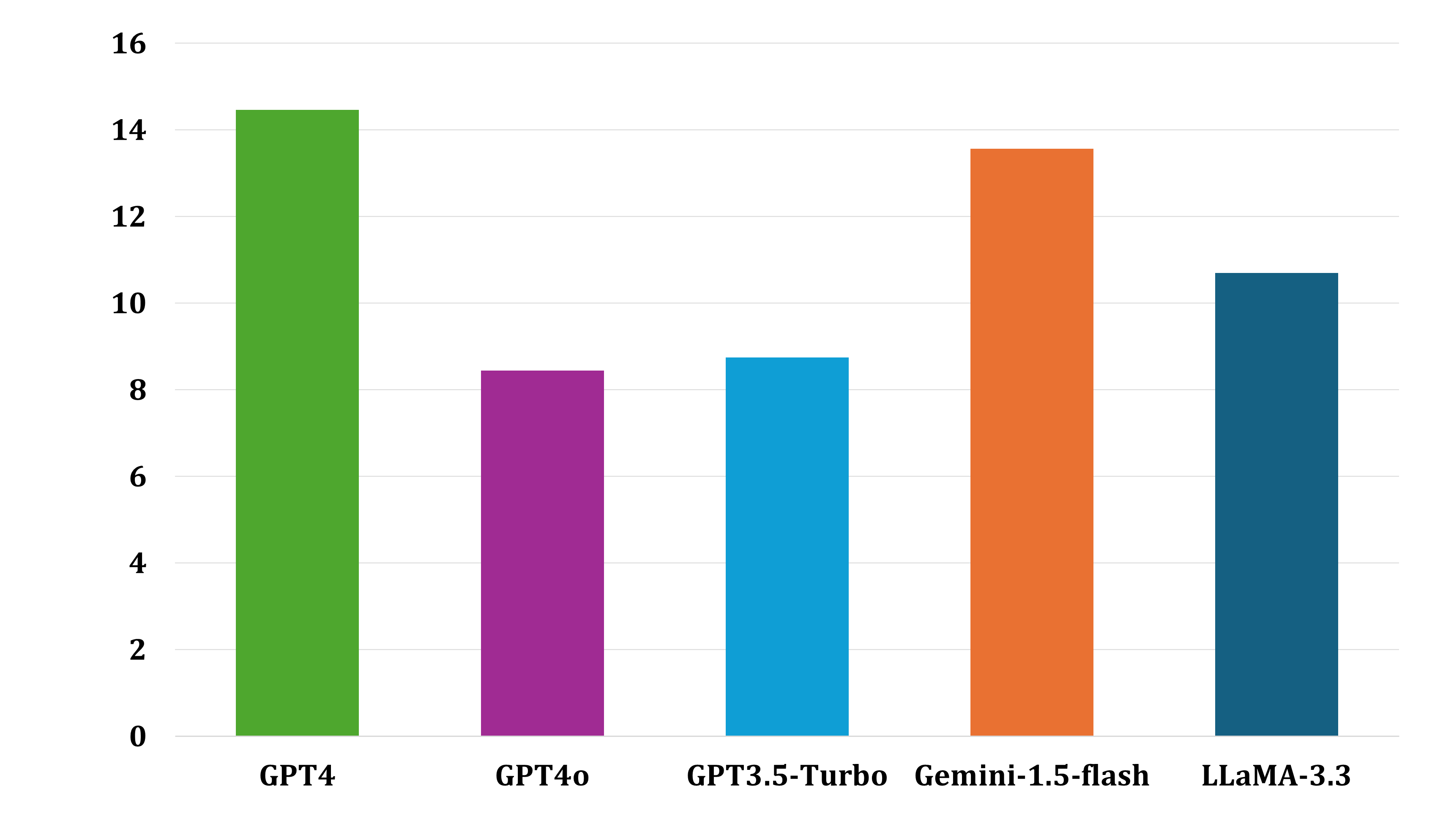}
    \caption{Comparison of average response time (in seconds) for different LLMs to generate the symbolic plan. GPT-4 exhibits the highest response time, while GPT-4o and GPT-3.5-Turbo are comparatively faster. Gemini-1.5-flash and LLaMA-3.3 provide intermediate performance, highlighting trade-offs between efficiency and model complexity.}
    \label{fig:llm-time}
\end{figure}

\begin{figure}[t]
    \centering
    \includegraphics[width=\linewidth]{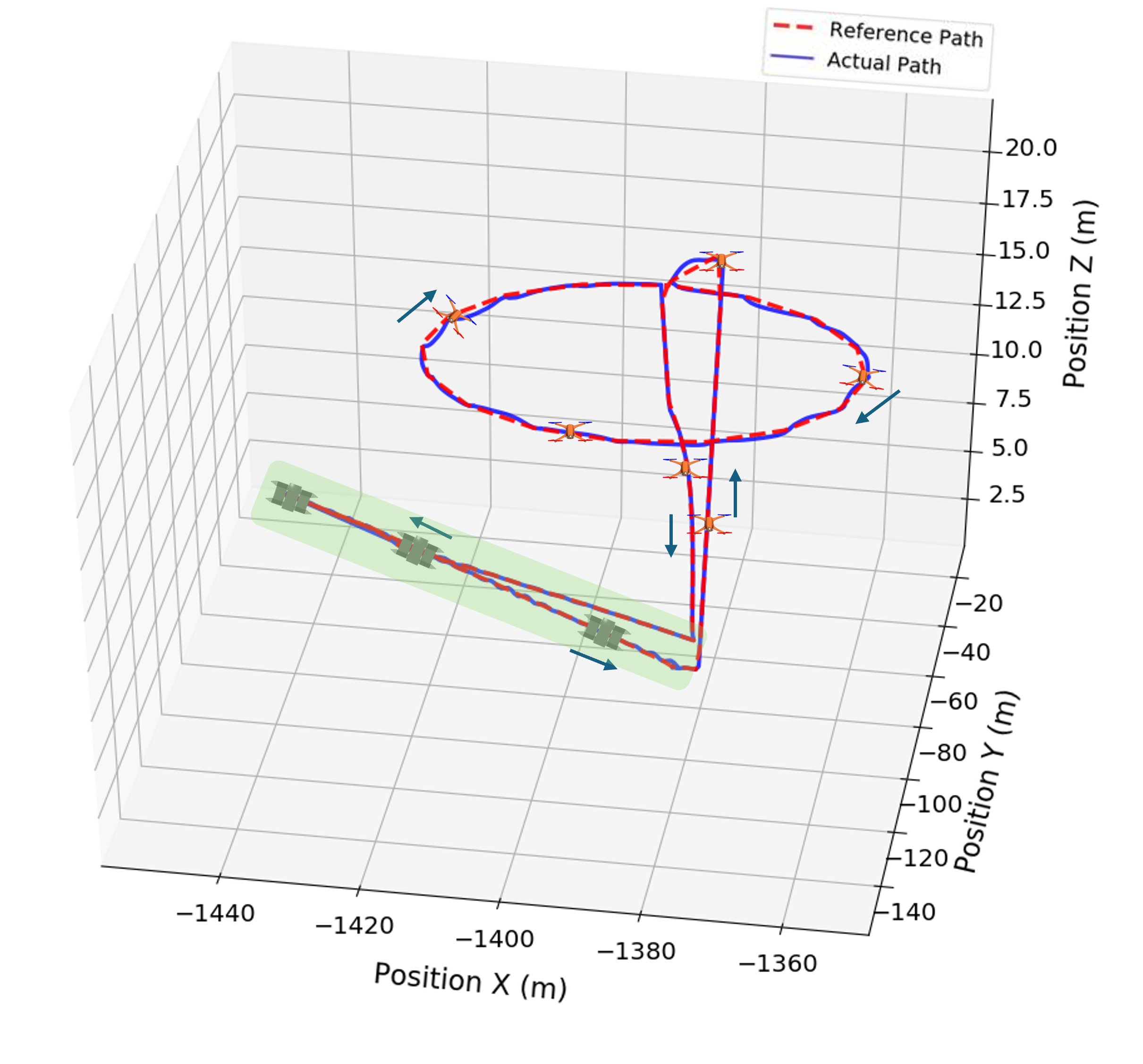}
    \caption{Reference and actual paths of the heterogeneous USV-UAV system during the crane inspection mission. The green rectangle highlights the USV trajectory while transporting the UAV to the deployment area. The remaining circular trajectory corresponds to the UAV path, which performs a 360 degree orbit around the crane for inspection. The comparison shows close tracking of the reference path by the actual execution of both platforms.}
    \label{fig:path-circular}
\end{figure}
\begin{figure}[t]
    \centering
    \includegraphics[width=\linewidth]{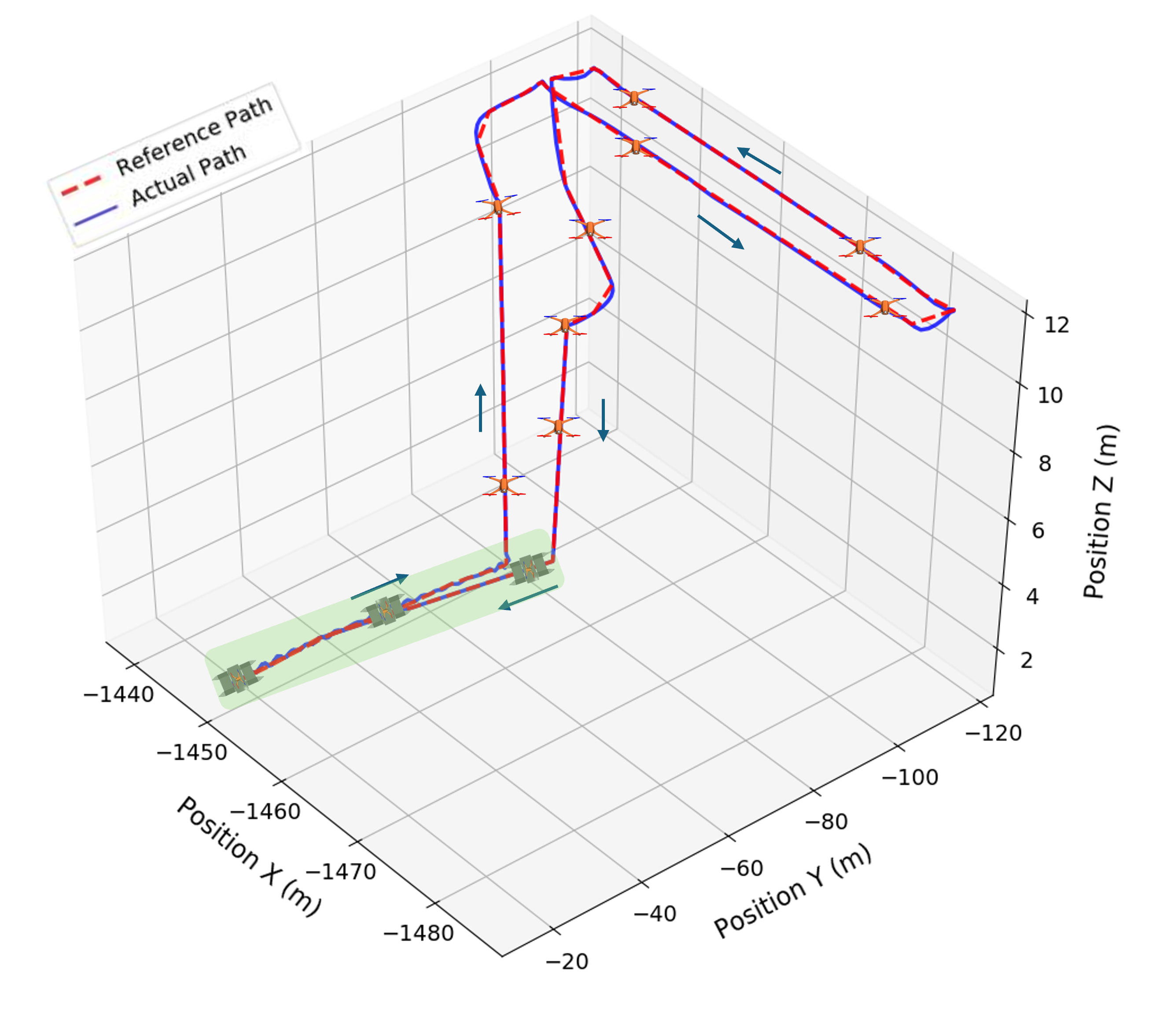}
    \caption{Reference and actual paths of the heterogeneous USV-UAV system during the container stack inspection mission. The green rectangle indicates the USV trajectory while transporting the UAV to the deployment point. The UAV then follows a rectangular trajectory around the container stacks to perform inspection, with the actual path closely matching the reference path.}
    \label{fig:path-rectangular}
\end{figure}
\begin{figure}[h]
    \centering
    \includegraphics[width=\linewidth]{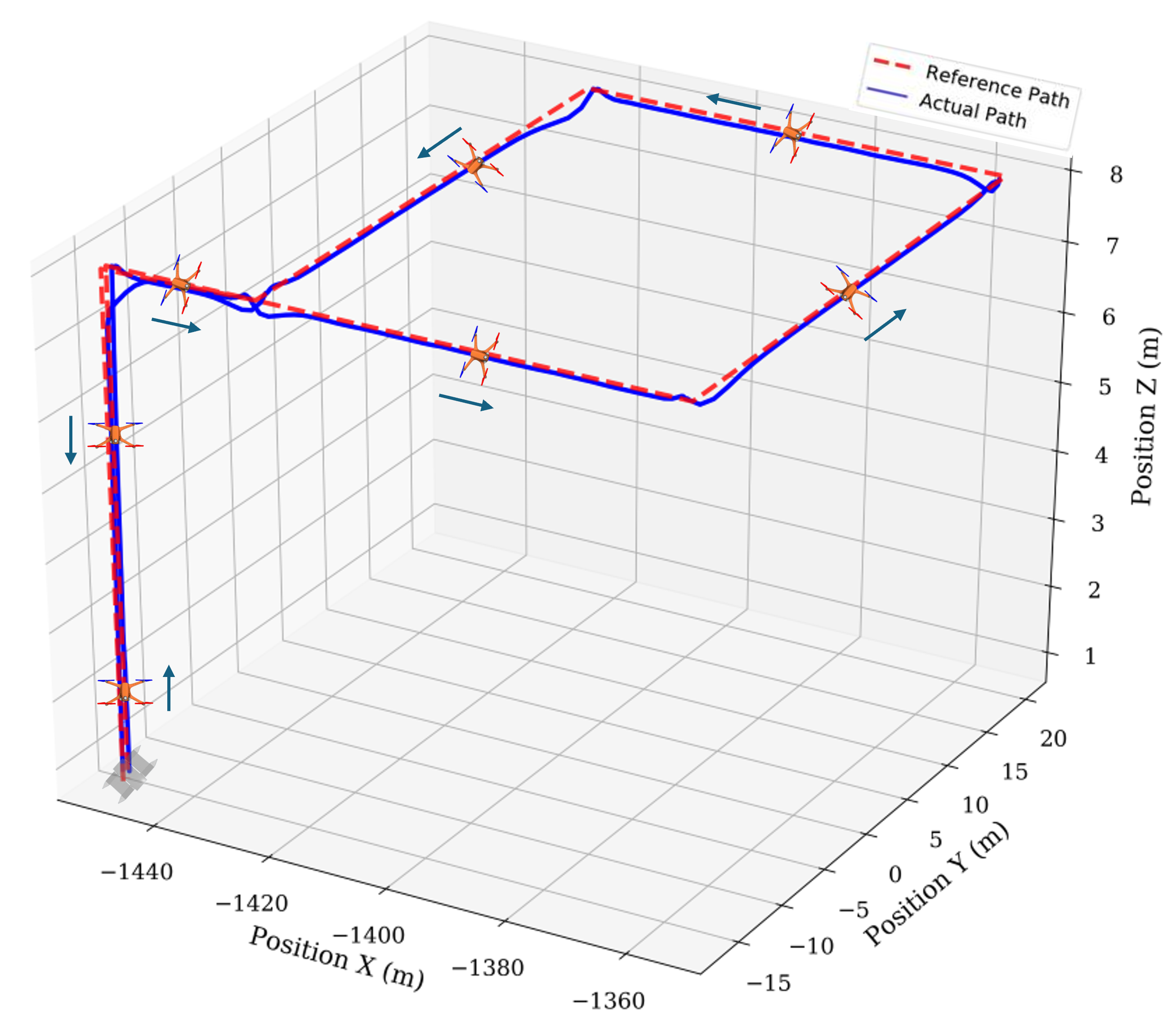}
    \caption{Reference and actual UAV paths during the central docking station inspection mission. In this scenario, only the UAV was deployed directly from the home location without USV repositioning. The UAV follows a rectangular trajectory around the docking station, with the actual flight path closely matching the reference path.}
    \label{fig:path-hour}
\end{figure}
The results in Table~\ref{table-llm} show different performance patterns between simple and complex tasks. For simpler missions (Tasks 1-3), which involved straightforward surveillance or waypoint following by a single vehicle, all models performed relatively well. GPT-4o, GPT-3.5-Turbo, and Gemini achieved correctness scores greater than 92 with execution success rates exceeding 80\%. However, LLaMA often produced invalid output, with correctness dropping below 50 and success rates close to 30\%.  

For more complex missions (Tasks 4-5), requiring simultaneous UAV-USV coordination, such as joint dockside inspections with USV and container stack inspection with UAV. GPT-4o maintained correctness above 94 and success rates around 85-87\%, with response times below 9 seconds. GPT-3.5-Turbo produced comparable correctness but slightly reduced success rates (around 80\%), although it responded faster. Gemini degraded more sharply in these tasks, with the correctness averaging 82 and execution success near 70\%. LLaMA failed to generate viable multiagent plans, remaining below 43 in correctness and 32\% in execution success.  

In general, GPT-4o demonstrated the best balance in correctness, execution success, and response time, confirming its suitability for time sensitive heterogeneous maritime missions. The expert driven evaluation highlights that both correctness and execution success are essential: correctness ensures plans are syntactically and semantically valid, while execution success validates their practical applicability in real-world UAV-USV coordination.

The bar plots in Fig.~\ref{fig:llm-success} and Fig.~\ref{fig:llm-time} illustrate the average performance of the evaluated LLMs across all maritime mission planning tasks.  Figure~\ref{fig:llm-success} compares \textit{Correctness} and \textit{Execution Success Rate}, showing that GPT-4o, GPT-3.5-Turbo, and GPT-4 consistently achieve higher correctness scores (above 90) and success rates (above 80\%), while Gemini shows moderate performance and LLaMA lags significantly behind. Figure~\ref{fig:llm-time} presents the average 
\textit{Response Time}, highlighting GPT-4o and GPT-3.5-Turbo as the fastest models (below 9 seconds), while GPT-4 and Gemini incur longer latencies. Together, these plots reinforce that GPT-4o provides the best overall balance of correctness, execution reliability, and response speed.  

Fig.~\ref{fig:path-circular},~\ref{fig:path-rectangular}, and~\ref{fig:path-hour} illustrate the trajectories followed by the USV and UAV during different inspection missions. In each case, the green rectangle denotes the path of the USV while carrying the UAV to its designated take-off location. Fig.~\ref{fig:path-circular}, shows the path corresponding to the mission \textit{Inspect unauthorized personnel/vehicles in the crane work zone}. The USV navigates close to the crane, after which the UAV takes off and performs an aerial inspection along a circular path around the crane. Fig.~\ref{fig:path-rectangular} presents the results of the mission \textit{Survey the container stack and check the stacking}. Here, the USV transports the UAV to a position near the container stacks, from which the UAV is deployed to inspect the area following a rectangular trajectory. Finally, Fig.~\ref{fig:path-hour} depicts the mission \textit{Inspect the central docking zone and detect unauthorized sailboats}. Since the docking station is located close to the home position, the UAV takes off directly, executes the inspection along a rectangular path, and returns without requiring USV repositioning.

\subsubsection{VLM-based Inspection Results}
For inspection missions, we adopted pre-trained lightweight VLMs that can be efficiently deployed on edge devices and embedded systems. Unlike large-scale foundation models that require substantial GPU clusters, these compact VLMs offer a practical balance between computational efficiency, memory footprint, and reasoning capability. Their lightweight design makes them particularly suitable for real-world maritime operations, where onboard processing operates under strict energy and resource constraints. Table~\ref{tab:vlm-mem} provides a summary of the VLMs benchmark for port inspection, including model identifier, approximate parameter size, supported device types, input resolution, and peak memory usage. The selection spans a spectrum of architectures, from SmolVLM with approximately 0.5B parameters and a modest memory requirement of 2.5 GB, to Qwen2-VL Instruct with 2B parameters, which despite higher resource demand, provides stronger reasoning capabilities. Intermediate models such as Florence-2, moondream2, and GIT-base cover diverse trade-offs between size, accuracy, and device compatibility.

\begin{table*}[t]
\centering
\caption{Lightweight VLMs evaluated for port inspection tasks, with details on model size, device compatibility, input resolution, and memory footprint. These models were selected for their suitability for on-device deployment in resource-constrained maritime inspection scenarios.}\label{tab:vlm-mem}
\begin{tabular}{|p{4.8cm}|p{2.5cm}|p{3cm}|p{3cm}|p{2.5cm}|}
\hline
\textbf{Model ID (Reference)} & \textbf{Approx. Params} & \textbf{Device} & \textbf{Resolution} & \textbf{Memory} \\
\hline
SmolVLM \cite{marafioti2025smolvlm} & \(\sim0.5\)B & CPU / GPU & \(384 \times 384\) & \(\sim\)2.5 GB \\ \hline
Florence-2 \cite{xiao2024florence2} & \(\sim230\)M & CPU / GPU & \(448 \times 448\) & \(\sim\)3.2 GB \\ \hline
moondream2 \cite{vikhyat2024moondream2} & \(\sim1.3\)B & GPU & \(378 \times 378\) & \(\sim\)4.1 GB \\ \hline
GIT-base  \cite{wang2022git} & \(\sim345\)M & CPU / GPU & \(224 \times 224\) & \(\sim\)1.8 GB \\ \hline
Qwen2-VL \cite{bai2024qwen2vl} & \(\sim2\)B & GPU / CPU & \(448 \times 448\) & \(\sim\)5.2 GB \\ \hline
\end{tabular}
\end{table*}
\begin{figure*}[!t]
    \centering
    \includegraphics[width=\linewidth]{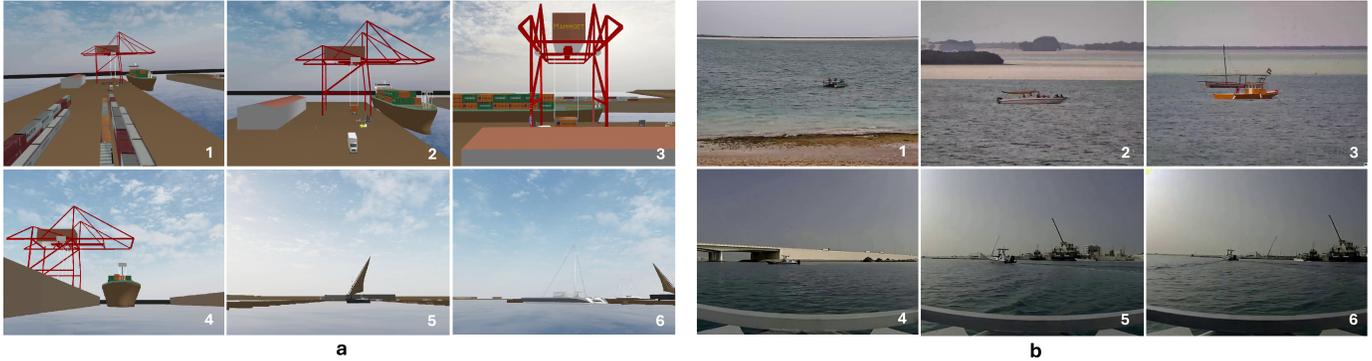}
    \caption{Example camera streams from simulation and real-world deployments. (a) Simulation environment: the first row shows UAV camera streams during inspection missions, while the second row shows USV camera streams. (b) Real robot camera stream: the first row illustrates UAV camera streams, and the second row depicts USV camera streams captured during port inspection tasks.}
    \label{fig:stream}
\end{figure*}
\begin{figure}[t]
    \centering
    \includegraphics[width=\linewidth]{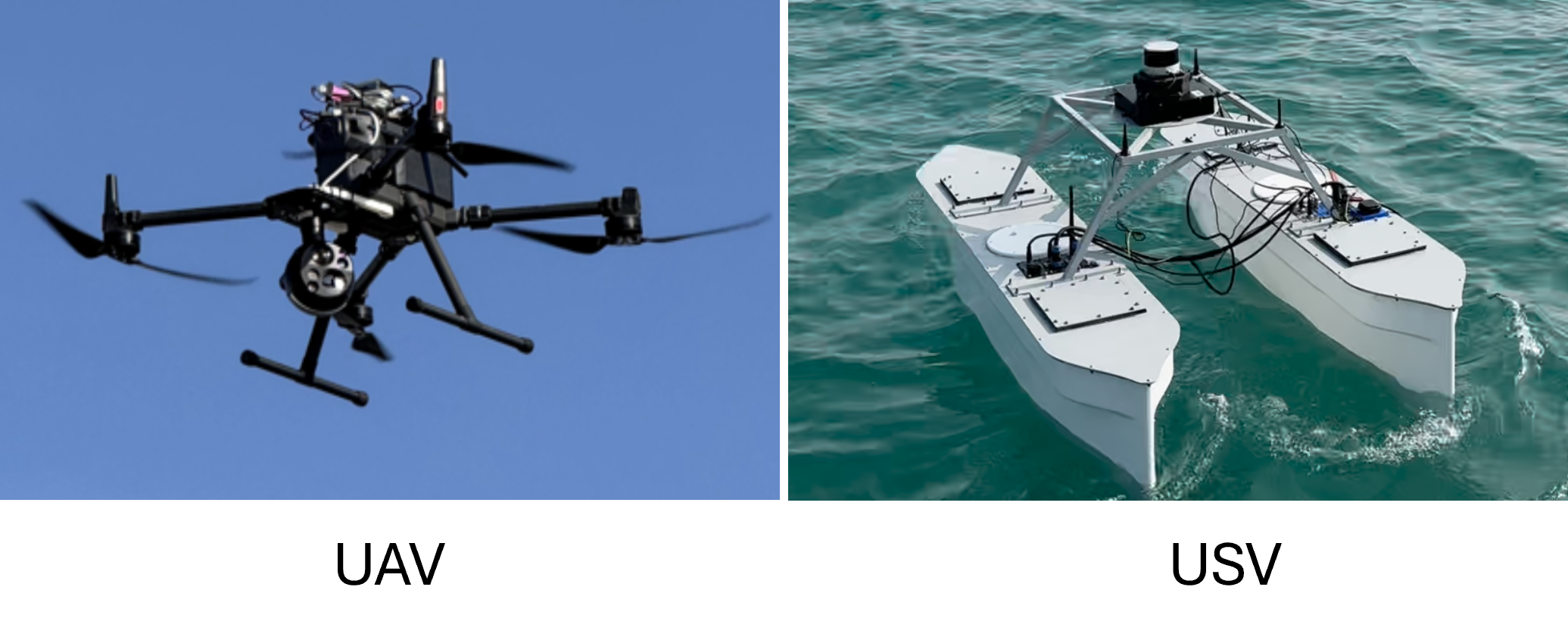}
    \caption{Real UAV (left) and USV  (right) platforms used to capture camera streams in real maritime conditions. These UAV and USV camera streams serve as input data for evaluating VLMs on real data.}
    \label{fig:real-uav-usv}
\end{figure}
To evaluate their utility for maritime security applications, we designed a set of domain-specific natural language queries to run on each model for every captured image. These queries were formulated to identify unusual activities or unauthorized access in different operational zones of the port. Representative examples include: \textit{Is there any human in the image near the crane?} to monitor restricted work areas, \textit{Is there any vehicle visible on the scene?} to detect unauthorized ground access, \textit{Is there a sailboat in the docking station?} to identify intrusions in critical docking zones, and \textit{Is there any boat in this port scene?} for general vessel detection in the environment. By systematically applying these structured queries across all benchmark models, we evaluated both their accuracy in recognizing objects and their capacity for semantic reasoning within specific maritime scenarios. This evaluation highlights how lightweight VLMs can serve as reliable on-device perception modules, providing real-time situational awareness and decision support without relying on cloud-based inference or heavy computational infrastructure.

The VLMs are evaluated using camera streams obtained from both simulation and real-world experiments with the UAV and USV platforms. Fig.~\ref{fig:stream} presents representative snapshots of camera streams captured by the UAV and USV in these two settings. Specifically, Fig.~\ref{fig:stream}a corresponds to the simulated camera streams, where the first row shows the UAV view and the second row shows the USV view. In contrast, Fig.~\ref{fig:stream}b illustrates the real-world camera streams, with the first row showing the UAV view and the second row showing the USV view. The actual UAV and USV platforms that are used to capture the real camera streams are depicted in Fig.~\ref{fig:real-uav-usv}.
\begin{figure*}[t]
    \centering
    \includegraphics[width=\linewidth]{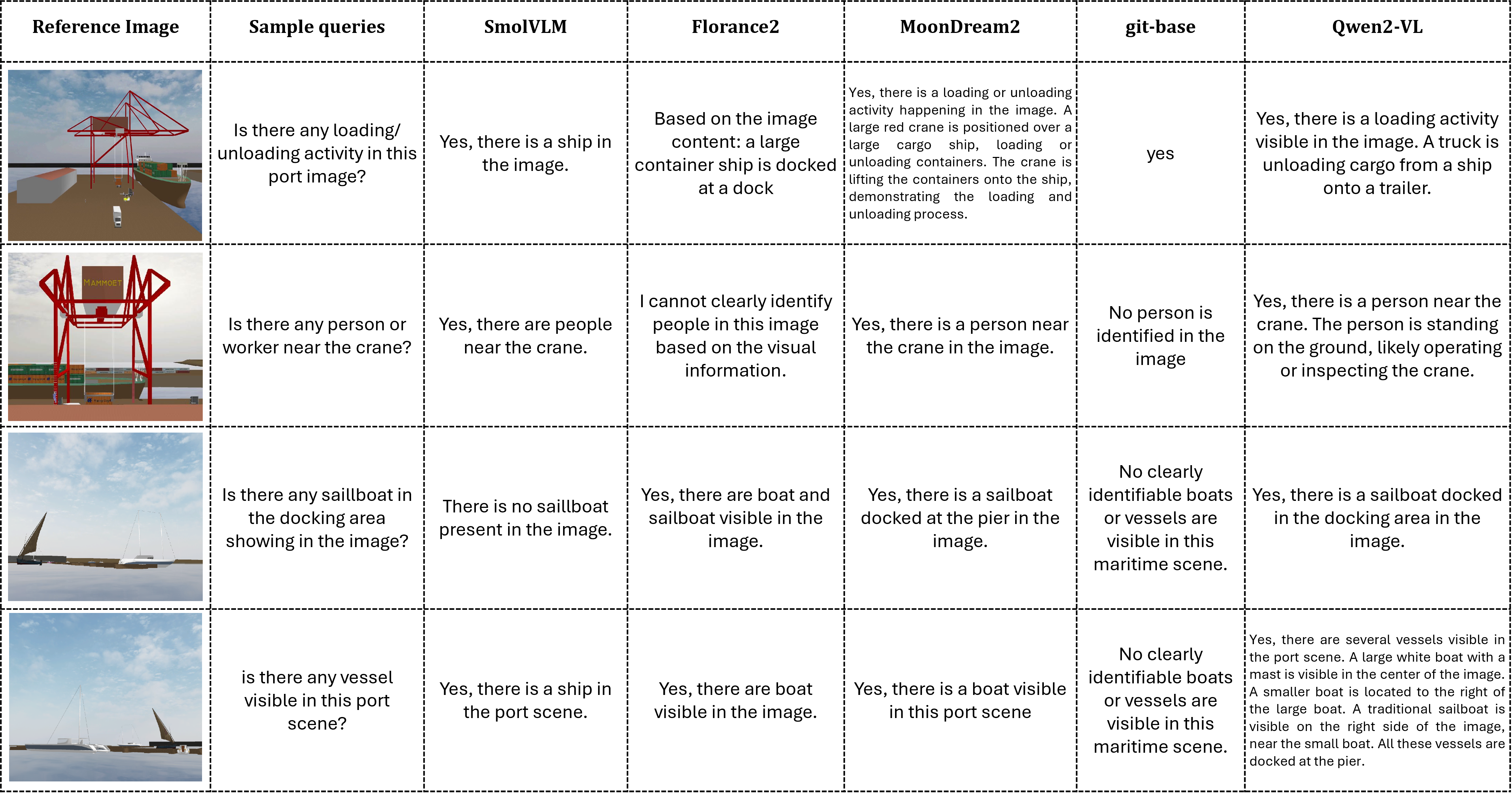}
    \caption{Sample responses of different VLMs when tested on UAV and USV camera streams obtained from the simulation environment. The table highlights how each model interprets domain-specific queries related to port inspection tasks such as detecting vessels, sailboats, workers, and loading/unloading activities.}
    \label{fig:sim-tab}
\end{figure*}

The results in Fig~\ref{fig:sim-tab} show the responses obtained when camera streams from either the UAV or the USV are fed to each pretrained VLMs. The queries were designed to investigate four representative inspection conditions: (i) detection of loading/unloading activity near a crane, (ii) presence of a person or worker in the scene, (iii) identification of sailboats in the docking area, and (iv) general identification of the vessel.
Across these scenarios, Qwen2-VL and Moondream2 consistently generated the most context-rich responses, often providing spatial cues such as \emp{near the crane} or \emp{on the pier}, which are particularly valuable for downstream evaluation and situational awareness. Florance2 produced compact but generally correct responses in the form of binary Yes/No judgments with brief justification. The SmolVLM and GIT-base were comparatively conservative, especially for challenging human detection, frequently returning uncertainty statements when the visual evidence was small, distant, or partially occluded.
In particular, the human presence query exposed a shared weakness across smaller models, leading to false negatives under low-resolution or distant viewpoints. In contrast, crane activity and general vessel visibility were comparatively easier tasks, with a wider agreement among models. Overall, these results highlight that (a) personnel detection at distance remains the most challenging simulation case, and (b) models with larger or better-tuned language heads (Qwen2-VL, Moondream2, Florence-2) produce more actionable and localized descriptions, while lighter models like SmolVLM and GIT-base struggle with complex or fine-grained perception.

\begin{figure*}[t]
    \centering
    \includegraphics[width=\linewidth]{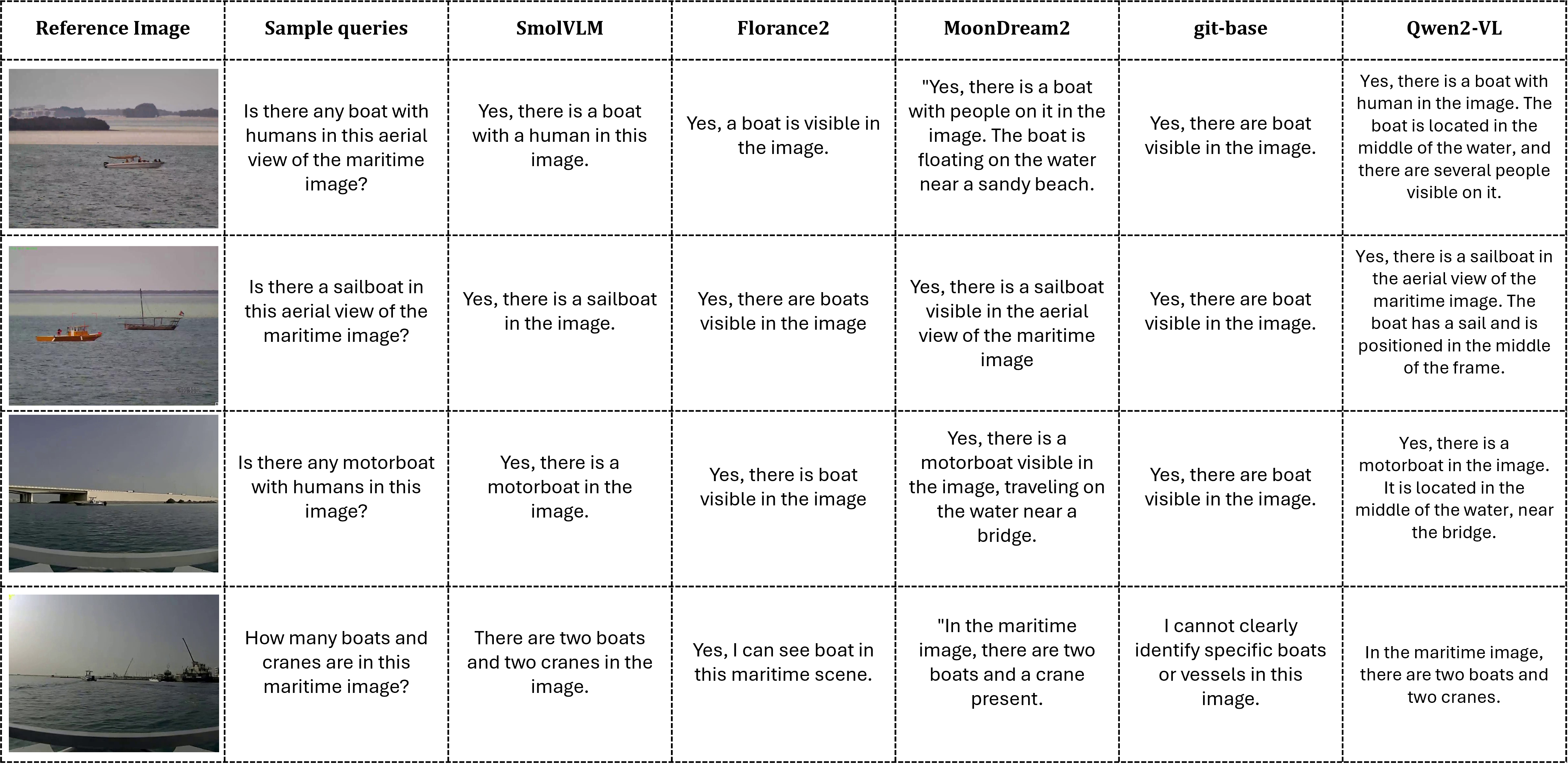}
    \caption{Responses of different VLMs when tested on real-world UAV and USV camera streams. The queries cover representative maritime inspection conditions including vessel detection, sailboat identification, personnel presence, and port activity recognition}
    \label{fig:real-tab}
\end{figure*}

The results in Fig.~\ref{fig:real-tab} summarize VLM performance when real UAV and USV camera streams were used as input. As in the simulation evaluation, queries were designed around four representative maritime inspection tasks: (i) vessel detection, (ii) identification of sailboats, (iii) recognition of personnel, and (iv) port activity observation.
In this real-world setting, Qwen2-VL and Moondream2 again produced the most detailed and contextually grounded responses, often localizing objects relative to landmarks (e.g., “near the pier”, “beside another boat”). SmolVLM showed stronger performance than in simulation, delivering concise but generally accurate responses across most queries, making it more reliable than Florence-2, which remained cautious and prone to uncertainty in fine-grained detections.
By contrast, the GIT-base was the weakest model, frequently producing incorrect or incomplete responses and particularly struggling with personnel and sailboat identification. Factors such as variable resolution, lighting conditions, and motion blur in the real environment further exposed these limitations. However, vessel detection tasks were comparatively simpler and resulted in greater agreement among most VLMs.
Overall, these real-environment experiments confirm the relative ranking of models: Qwen2-VL and Moondream2 provide the most actionable and context-rich output, SmolVLM offers competitive performance for its size, Florence-2 is more conservative, and GIT-base lags significantly behind.

The performance of the VLM models across images captured by UAV and USV in simulation and in real environment is assessed through semantic correctness (SC).  It was evaluated using a two step expert review process. First, the predictions of each VLM were compared with ground-truth annotations prepared by domain experts with experience in maritime robotics. Each answer was scored on three levels: $0$ indicated an incorrect answer, $0.5$ represented a partially correct answer (for example, an answer that contained "yes" but lacked detail or provided vague descriptions), and $1$ denoted a correct answer with semantic detail, including contextual or spatial reasoning. Two independent experts scored each trial and disagreements were resolved through discussion to ensure consistency. The final scores reflect the average of all the annotators. This procedure ensured that the semantic correctness metric captured not only the detection accuracy but also the quality of contextual reasoning, which is critical for maritime inspection scenarios.

\begin{table}[t]
\centering
\caption{Performance of lightweight VLMs across UAV and USV  in simulation and in real maritime environment SC: indicated the semantic correctness, IT: is the average inference time in seconds.}
\label{tab:sc}
\renewcommand{\arraystretch}{1.2}
\setlength{\tabcolsep}{2pt}
\begin{tabularx}{\columnwidth}{
  >{\hsize=0.3\hsize}X   
  |>{\hsize=0.35\hsize}X 
  |>{\hsize=0.2\hsize}X  
  |>{\hsize=0.15\hsize}X 
}
\hline
\textbf{Model} & \textbf{Scenario} & \textbf{SC (\%)} & \textbf{IT (s)} \\
\hline
\multirow{4}{*}{SmolVLM} 
& UAV-Real & 78.0 & 0.353 \\
& USV-Real & 80.1 & 0.292 \\
& UAV-Sim  & 65.0 & 0.505 \\
& USV-Sim  & 62.4 & 0.572 \\
\hline
\multirow{4}{*}{Florence-2} 
& UAV-Real & 52.6 & 0.324 \\
& USV-Real & 48.0 & 0.310 \\
& UAV-Sim  & 66.7 & 0.331 \\
& USV-Sim  & 75.0 & 0.327 \\
\hline
\multirow{4}{*}{Moondream2} 
& UAV-Real & 80.4 & 0.299 \\
& USV-Real & 83.3 & 0.258 \\
& UAV-Sim  & 82.5 & 0.314 \\
& USV-Sim  & 81.7 & 0.295 \\
\hline
\multirow{4}{*}{GIT-base} 
& UAV-Real & 42.0 & 0.346 \\
& USV-Real & 38.0 & 0.344 \\
& UAV-Sim  & 50.0 & 0.331 \\
& USV-Sim  & 47.4 & 0.344 \\
\hline
\multirow{4}{*}{Qwen2-VL} 
& UAV-Real & 82.7 & 0.592 \\
& USV-Real & 84.5 & 0.566 \\
& UAV-Sim  & 83.3 & 0.624 \\
& USV-Sim  & 83.3 & 0.589 \\
\hline
\end{tabularx}
\end{table}

The results in Table~\ref{tab:sc} highlight different performance patterns between models. Qwen2-VL achieved the highest overall semantic correctness, consistently exceeding 82\% in both UAV and USV scenarios, with only modest inference time overhead (~0.6s). Moondream2 also demonstrated strong semantic reasoning, maintaining SC scores above 80\% in both simulation and real deployments, while providing faster responses than Qwen2-VL. SmolVLM delivered competitive performance in real world UAV/USV settings (~80\%) but showed a drop in simulation, suggesting a less robust generalization. Florence-2 produced mixed results, performing better under simulated conditions (up to 75\%) but falling below 55\% in real maritime conditions, indicating sensitivity to real world noise. GIT-base consistently underperformed, with SC scores below 50\%, confirming its limitations for semantic maritime inspection tasks. Overall, the comparison shows that while all lightweight models can operate with subsecond inference times, Qwen2-VL and Moondream2 strike the best balance between semantic reasoning and efficiency, making them the most promising candidates for on-board deployment in USV-UAV cooperative inspection missions.
\begin{figure}[t]
    \centering
    \includegraphics[width=\linewidth]{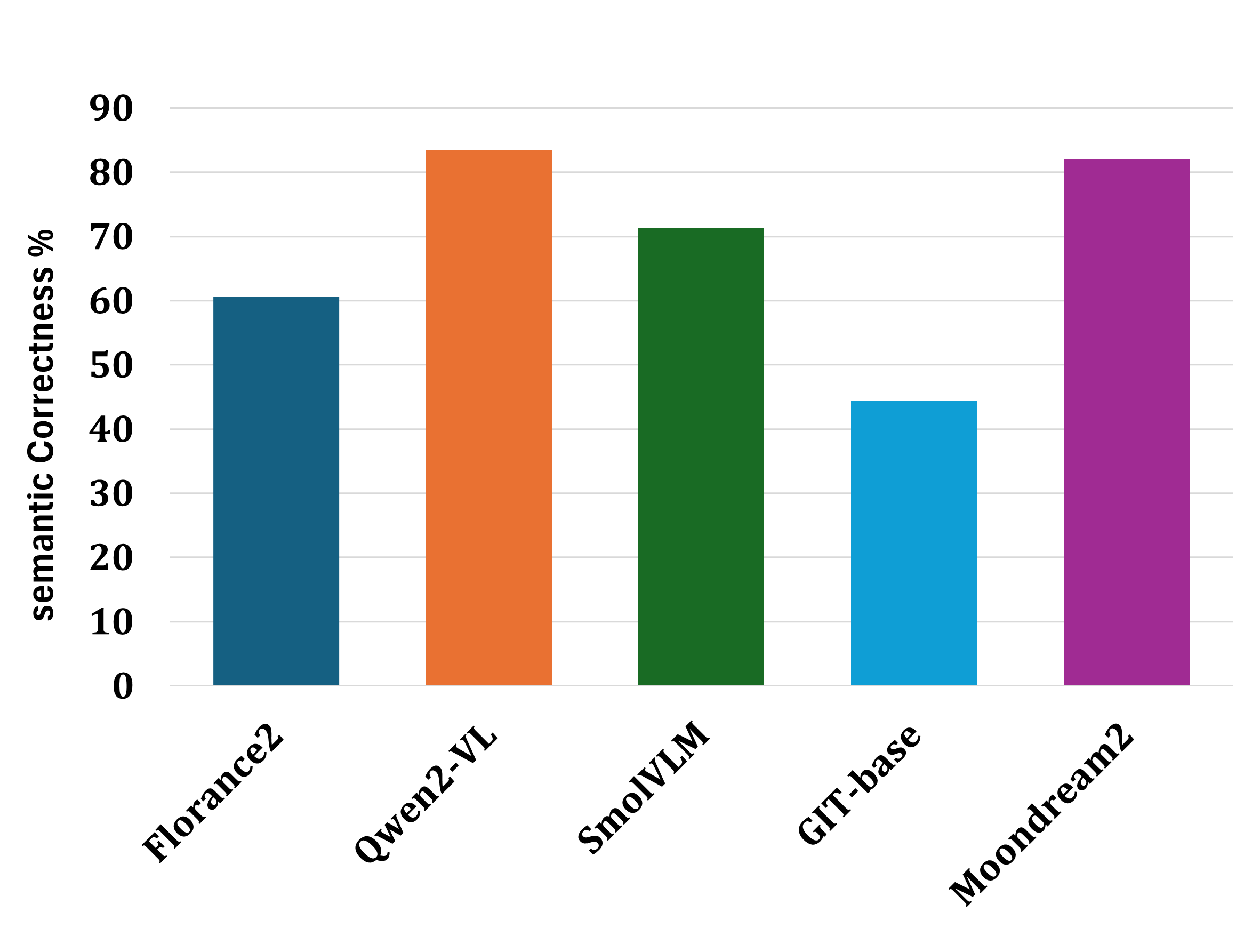}
    \caption{Average semantic correctness (\%) of lightweight VLMs evaluated across UAV and USV inspection tasks in both simulation and real maritime conditions.}
    \label{fig:cor}
\end{figure}
 
\begin{figure}[t]
    \centering
    \includegraphics[width=\linewidth]{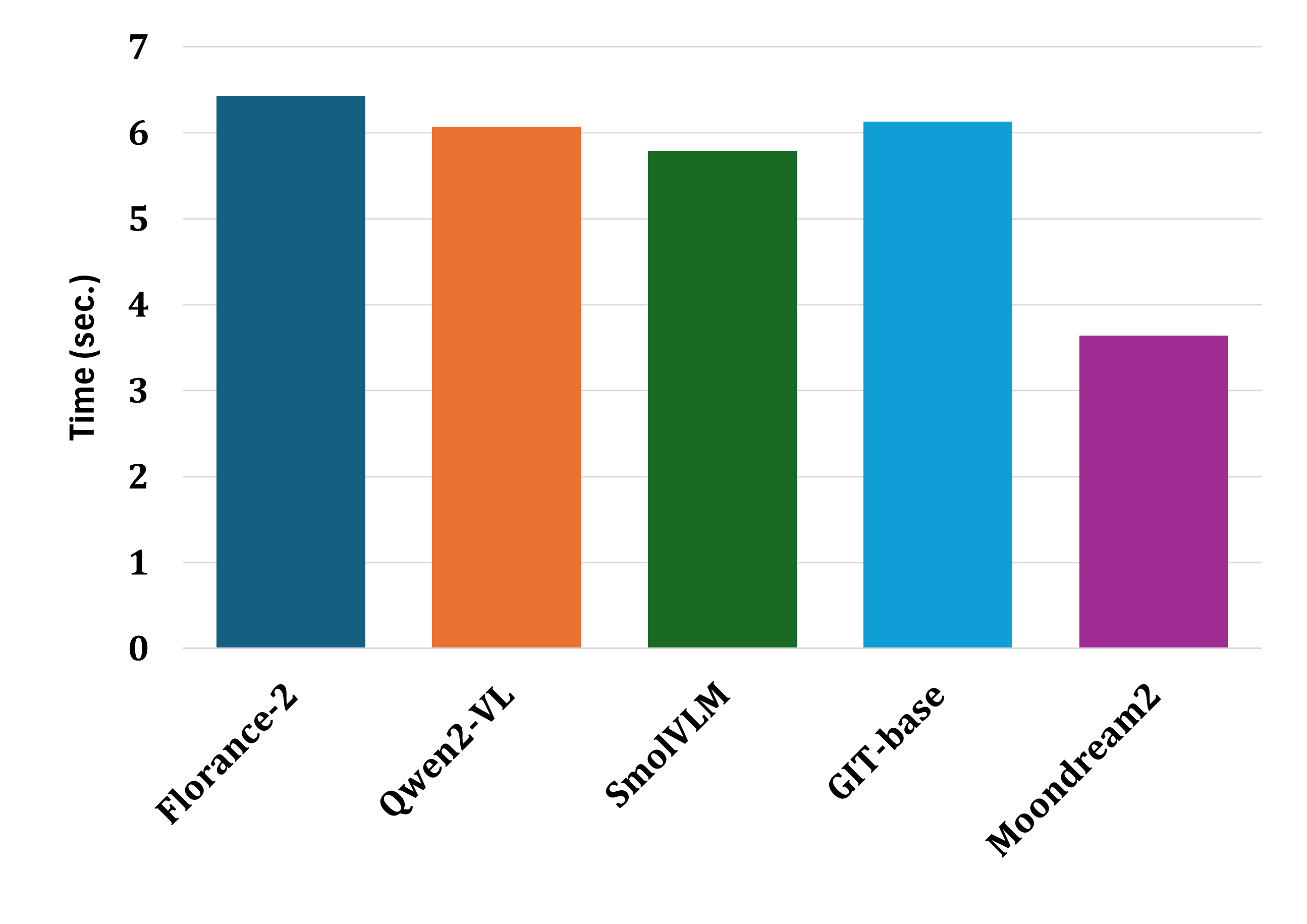}
    \caption{Model loading times of lightweight VLMs benchmark for port inspection tasks. The results show variation in initialization overhead across models, with Moondream2 demonstrating the fastest loading time, while Florence-2 requires the longest startup.}
    \label{fig:vlm-loading}
\end{figure}
\begin{figure}[t]
    \centering
    \includegraphics[width=\linewidth]{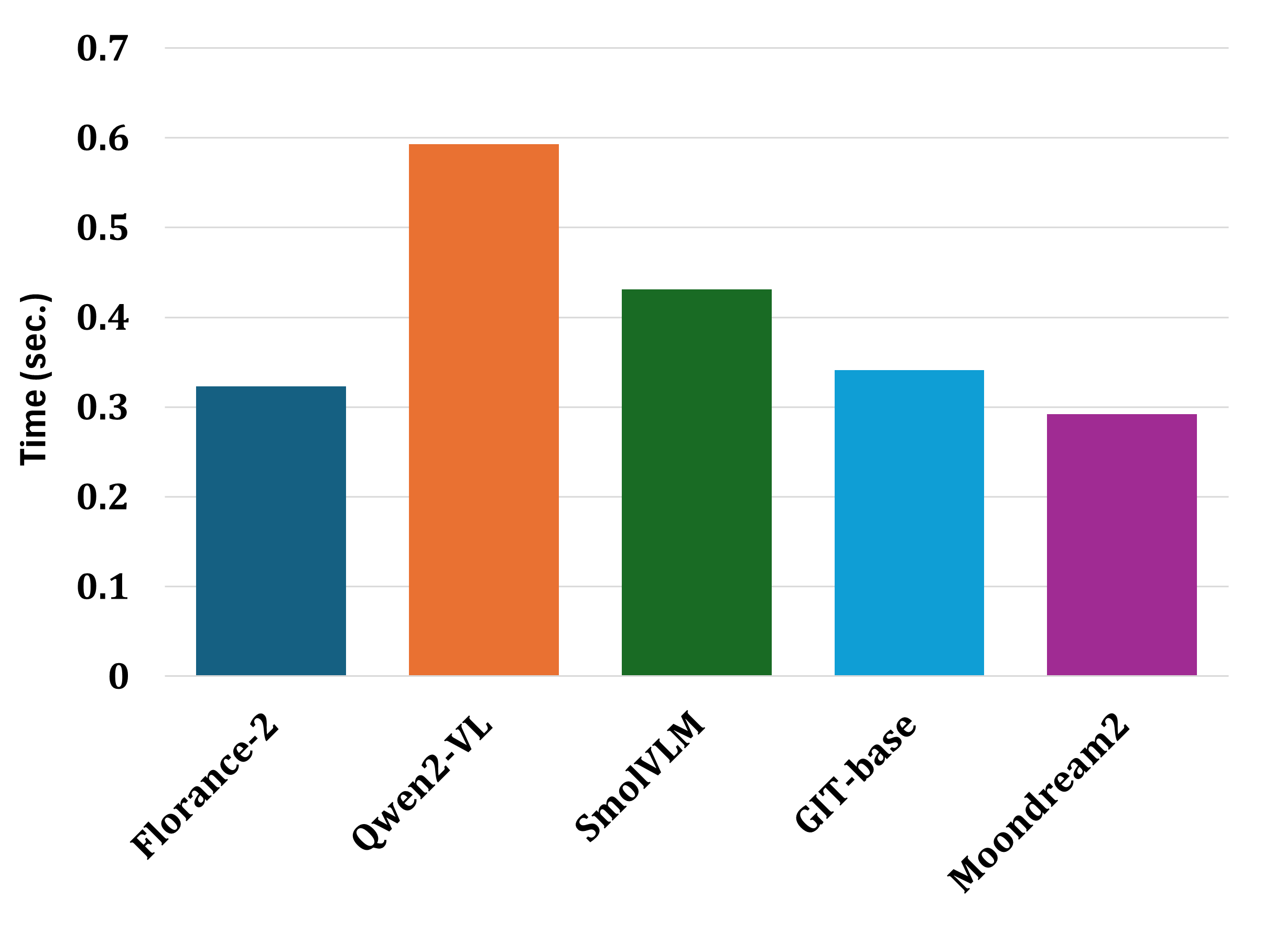}
    \caption{Inference times of lightweight VLMs for a single query on inspection images. GIT-base achieves the lowest response latency, whereas Florence-2 shows the highest, reflecting differences in model size and computational efficiency.}
    \label{fig:vlm-inference}
\end{figure}
The overall average semantic correctness score for each model is shown in Fig.~\ref{fig:cor}.  Qwen2-VL achieved the best overall performance with an average correctness above 83\%, closely followed by Moondream2 with 82\%. SmolVLM also performed competitively, maintaining a correctness of over 70\%. In contrast, Florence-2 struggled to generalize effectively, achieving only around 60\%, while GIT-base consistently underperformed at less than 45\%. These findings suggest that while most modern lightweight VLMs are capable of semantic reasoning in maritime inspection tasks, only a subset (Qwen2-VL, Moondream2, and SmolVLM) demonstrate sufficient robustness for practical deployment.

In addition to semantic correctness, we evaluated the computational efficiency of each VLM in terms of average inference time (Fig.~\ref{fig:vlm-inference}). The results show a clear trade-off between model size and speed. Florence-2 and GIT-base achieve relatively low latency, averaging around 0.32 to 0.34 seconds, while SmolVLM takes slightly longer at 0.43 seconds. Moondream2 stands out as the most efficient model with the lowest inference time (0.29 seconds), which makes it attractive for real-time on-device deployment. In contrast, Qwen2-VL, despite achieving the highest semantic correctness, is the slowest among the evaluated models at 0.59 seconds per query.

 we evaluated the computational efficiency of each VLM in terms of average loading time (Fig.~\ref{fig:vlm-loading}) and average inference time (Fig.~\ref{fig:vlm-inference}. The results highlight significant trade-offs between model complexity and runtime efficiency. For loading, Moondream2 clearly outperformed all others, initializing in only 3.6 s, compared to 6.4 s for Florence-2 and ~6.0 s for Qwen2-VL and GIT-base. SmolVLM fell in between with 5.8 s. Once loaded, the inference times were substantially lower, ranging between 0.29 and 0.59 s. Here, Moondream2 again demonstrated strong efficiency with ~0.29 s per query, closely followed by GIT-base (~0.34 s). Qwen2-VL was the slowest, averaging ~0.59 s, while Florence-2 and SmolVLM balanced moderate latency (0.32–0.43 s).
Together, these results demonstrate a performance efficiency divide: larger models like Qwen2-VL provide stronger semantic reasoning but generate higher latency, whereas lighter architectures like GIT-base are significantly more responsive but not provide high semantic accuracy. Models like Moondream2 are high in semantic correctness and have lower latency.

\section{Conclusions and Future Work}
This paper presented an integrated LLM-VLM fusion framework for autonomous maritime port inspection using cooperative UAV-USV systems. The framework addresses key limitations of conventional inspection methods by replacing rule-based mission planners with an LLM-driven symbolic planner and enhancing perception through VLM-based semantic inspection. The LLM component enables translation of high-level mission instructions into executable, dependency-aware task plans, ensuring safe and coordinated operation between aerial and surface platforms. The VLM module provides semantic scene understanding beyond predefined object detection, supporting context-aware anomaly detection, regulatory compliance assessment, and structured inspection reporting.
The proposed framework was validated through comprehensive simulations in the MBZIRC Maritime Simulator and verified using real-world inspection trials, demonstrating its reliability, adaptability, and suitability for deployment on resource constrained maritime platforms. In general, this study highlights the engineering potential of combining large language and vision language models to achieve semantically grounded, scalable, and autonomous maritime inspection. The proposed approach establishes a foundation for future intelligent marine operations, including vessel monitoring, offshore infrastructure inspection, and environmental surveillance.

Future research will focus on extending the proposed LLM-VLM fusion framework toward broader autonomous maritime applications and improved system efficiency. In particular, upcoming work will explore adaptive mission planning under dynamic environmental conditions, enabling real-time reconfiguration of UAV-USV coordination in response to changing sea states or inspection priorities. Lightweight optimization and model compression techniques will be investigated to further reduce computational load and enhance on-board inference performance for resource constrained robotic platforms.
Another important direction involves integrating additional sensing modalities such as hyperspectral imaging to enrich multimodal perception and improve semantic understanding of complex port and offshore structures. The framework will also be adapted for related maritime operations, including vessel monitoring, underwater infrastructure inspection, and oil spill detection. Finally, efforts will be made to evaluate long-term system reliability and safety in real sea trials, contributing to the development of fully autonomous, AI-driven maritime inspection and monitoring systems.
\section*{Acknowledgment}
\noindent This work was supported by Khalifa University under Award Nos. KU-BIT-Joint-Lab-8434000534, RC1-2018-KUCARS-8474000136, CIRA-2021-085, MBZIRC-8434000194, KU-Stanford-8474000605, and 8475000016 (in collaboration with Dubai Future Labs for the project ``Underwater Maritime Monitoring'').
\section*{Declaration}
Declaration of generative AI and AI-assisted technologies in the writing process.
During the preparation of this work, the author(s) used ChatGPT to refine the English. After using this tool/service, the author(s) reviewed and edited the content as needed and take(s) full responsibility for the content of the publication.
\section*{ Data Availability Statement}
The authors confirm that the data supporting the findings of this study are available in the article [and/or] its supplementary materials at \textcolor{blue}{\url{https://github.com/Muhayyuddin/llm-vlm-fusion-port-inspection}}. The complete data that was used to generate the benchmarks will be made public in the same repository upon acceptance. 
\balance

\bibliographystyle{plainnat}
\bibliography{refs}

\end{document}